\documentclass[pdflatex,sn-nature]{sn-jnl}

\usepackage[T1]{fontenc}%
\usepackage{graphicx}%
\usepackage{subcaption}%
\usepackage{float}%
\usepackage{placeins}%
\usepackage{multirow}%
\usepackage{amsmath,amssymb,amsfonts}%
\usepackage{amsthm}%
\usepackage{mathrsfs}%
\usepackage[title]{appendix}%
\usepackage{xcolor}%
\usepackage{textcomp}%
\usepackage{manyfoot}%
\usepackage{booktabs}%
\usepackage{algorithm}%
\usepackage{algorithmicx}%
\usepackage{algpseudocode}%
\usepackage{listings}%
\usepackage{fancyhdr}

\usepackage{fancyhdr}

\fancypagestyle{topnumber}{
    \fancyhf{}
    \fancyhead[R]{\thepage}

}

\theoremstyle{thmstyleone}%
\theoremstyle{thmstyletwo}%

\theoremstyle{thmstylethree}%

\begin{document}

\title[Article Title]{NeuroVolve: Evolving personalized visual stimuli toward user-defined brain response targets}


\author[1,2]{\fnm{Haomiao} \sur{Chen}}\email{hc872@cornell.edu}

\author[4]{\fnm{Keith} \sur{W. Jamison}}\email{kwj2001@med.cornell.edu}

\author[1]{\fnm{Daniel} \sur{Chong}}\email{djc473@cornell.edu}

\author[1]{\fnm{Yunrui} \sur{Zhang}}\email{yz3275@cornell.edu}

\author[1,2,4]{\fnm{Mert} \sur{R. Sabuncu}}\email{msabuncu@cornell.edu}

\author*[3,4]{\fnm{Amy} \sur{Kuceyeski}}\email{amk2012@med.cornell.edu}

\affil[1]{\orgdiv{School of Electrical and Computer Engineering}, \orgname{Cornell University}, \orgaddress{\city{Ithaca}, \state{NY}, \country{USA}}}

\affil[2]{\orgdiv{School of Electrical and Computer Engineering}, \orgname{Cornell Tech}, \orgaddress{\city{New York}, \state{NY}, \country{USA}}}

\affil[3]{\orgdiv{Computational Biology}, \orgname{Cornell University}, \orgaddress{\city{Ithaca}, \state{NY}, \country{USA}}}

\affil[4]{\orgdiv{Department of Radiology}, \orgname{Weill Cornell Medicine}, \orgaddress{\city{New York}, \state{NY}, \country{USA}}}


\abstract{A central question in neuroscience is what visual features drive brain responses across regions. Classic approaches rely on handcrafted stimuli to test specific hypotheses, limiting discovery to features defined before experiments. Here we introduce NeuroVolve, a data-driven generative framework for synthesizing images that achieve flexible, user-defined patterns of brain activation in individuals. NeuroVolve recovers established selectivity across visual cortical regions and, paired with subject-specific brain encoding models, generates personalized stimuli reflecting individual differences in neural preference. Beyond single-region selectivity, NeuroVolve uncovers multi-region response selectivity, characterizing how regions jointly represent visual content during naturalistic perception. We further demonstrate generalization beyond visual cortex: in a prospective fMRI experiment, NeuroVolve-generated stimuli targeting the amygdala—a classically non-sensory region—drove measured activation consistent with predictions in novel participants. Together, these results establish NeuroVolve as a less constrained, data-driven tool for probing individual-level neural selectivity across the brain.}




\maketitle
\pagestyle{topnumber}      
\thispagestyle{topnumber}  

The discovery that neurons respond preferentially to some 
stimuli over others \cite{hubel1962receptive, hubel1968receptive} 
has shaped our understanding of how the brain processes and 
interprets visual information. Over the past few decades, vision neuroscience researchers have conducted hypothesis-driven experiments using handcrafted stimuli, working under the assumption that specific cortical regions encode specific visual features. This 
approach has led to important insights into the human visual cortex. Early visual areas such as V1 correspond to low-level features such as orientation, spatial and temporal frequency, and
motion \cite{deangelis1995receptive, devalois1990spatial}. Recently, intermediate visual areas, like V2 or V4, were found to be responsive
to textures, curved contours or shapes\cite{nandy2013fine, ziemba2016selectivity}. While higher-order areas are selectively activated by semantically meaningful categories 
such as faces, bodies, places, words and, more recently, food \cite{grill2004human, kanwisher1997fusiform, downing2001cortical, sergent1992functional, Ishai1999, epstein1998cortical, jain2023selectivity, khosla2022highly}. Despite the successes of studies of this kind, a hypothesis-driven approach relies on features the experimenter has anticipated. Human bias likely mean missed stimulus properties that reflect the true "tuning" of brain areas. In less-studied regions, where strong prior hypotheses about driving features do not yet exist, hypothesis-driven stimulus design has limited reach. In an expansion of these focused, hypothesis driven experiments, the Natural Scenes Dataset (NSD) and others like it \cite{allen2022massive, chang2019bold5000} collected large-scale brain activity data in several individuals to measure their brain responses to thousands of images containing natural scenes, including outdoor and indoor scenes, animals, people, and food. These experiments paved the way for more unrestricted, data-driven methods of vision neuroscience discovery \cite{gu2022neurogen, khosla2022highly, luo2023brain, ponce2019evolving, iclrhmc}. 

Recent advances have leveraged AI-based image generative models and 
\emph{in silico} brain encoding models to characterize neural selectivity in a less restricted and data-driven manner \cite{gu2022neurogen, luo2023brain, 
ratan2021computational, ponce2019evolving, luo2023brainscuba, 
jang2025amygdala, bashivan2019neural, walker2019inception}.
Early approaches optimized image pixels using gradients from neural
encoding models, demonstrating effective neural stimulation in mice
and macaques \cite{bashivan2019neural,walker2019inception}. Subsequent methods constrained optimization through generative model priors: GANs \cite{goodfellow2020generative} in \cite{gu2022neurogen, ratan2021computational, jang2025amygdala} and diffusion models \cite{song2020score, ho2022cascaded} in \cite{luo2023brain}. These approaches produce more naturalistic images targeting maximal responses in particular parts of the human visual cortex. However, these procedures backpropagate through the full generator at every optimization step, tying optimization to the cost of the image generator. Furthermore, most of this body of work has focused on visual cortex; extending data-driven neural selectivity discovery to non-sensory regions, where semantic feature drivers are less well established, remains largely open.

\begin{figure*}
    \centering
    \includegraphics[width=\linewidth]{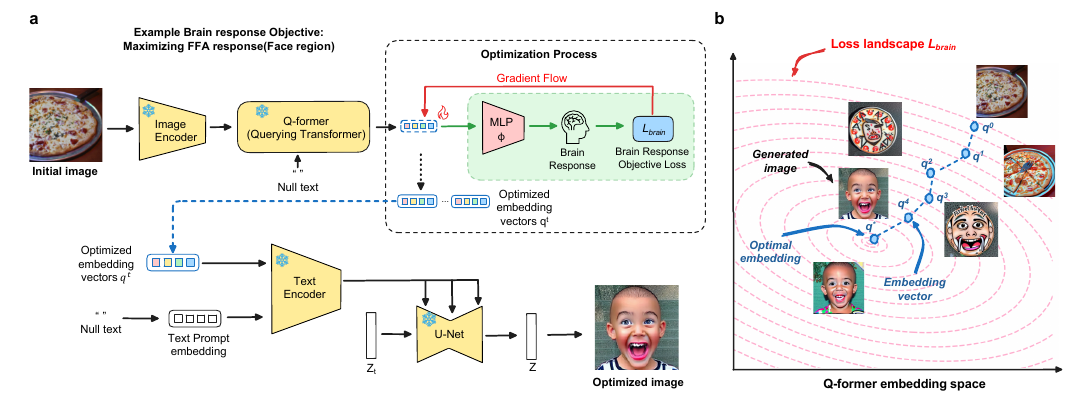}
   \caption{\textbf{\textit{NeuroVolve} framework for brain-response-guided image generation.}
\textbf{a}, Schematic of the NeuroVolve pipeline. An initial image is encoded into Q-Former (Querying Transformer) embeddings by a frozen image encoder and Q-Former. A pretrained subject-specific neural encoding model, $\Phi$, predicts voxelwise brain responses from these embeddings, and a user-defined brain-response objective guides their iterative optimization. The resulting optimized embeddings, $q^*$, are combined with a null-text prompt and used to condition the frozen BLIP-Diffusion model for image generation. Snowflake symbols denote frozen model components while fire symbols denote trained model components. The illustrated objective maximizes the predicted response of the fusiform face area (FFA). \textbf{b}, Conceptual illustration of the corresponding optimization trajectory in Q-Former embedding space. Pink dashed contours represent the brain-response objective loss landscape; blue points and arrows indicate successive embedding states from the initial embedding to the optimized embedding. Example images depict the visual content generated at selected points along the trajectory.}
    \label{fig:overview}
\end{figure*}

Building on these advances, we introduce NeuroVolve, a framework that reformulates neural selectivity characterization as a optimization problem in the image embedding space. Rather than testing pre-specified hypotheses one at a 
time, NeuroVolve "searches" for stimuli that achieve user-defined target 
patterns of brain activation, with a voxelwise encoder scoring 
each candidate by its predicted neural response. The search 
operates directly in the latent embedding space of a pretrained 
vision-language model — a rich learned representation of visual content — rather than through a specific image generator, enabling efficient optimization. Objectives are defined in the voxel-level 
activation space, so any differentiable combination of regions or voxels can serve as the optimization target, including single-region maximization and/or suppression and
more complicated multi-region objectives with arbitrary target activation levels. Optimized embeddings are 
decoded into coherent images by a frozen diffusion model. 
Fig.~\ref{fig:overview} provides an overview of the approach.

\section*{Results}

We demonstrate NeuroVolve's value through four contributions. 
First, NeuroVolve recovers established selectivity in early 
visual areas and category-selective regions of higher visual 
cortex, validating the data-driven approach against decades of 
hypothesis-driven findings. Second, paired with subject-specific 
brain encoders, NeuroVolve generates personalized stimuli that reflect 
individual differences in neural preference. Third, NeuroVolve synthesizes images that achieve multi-region brain-response objective — co-activation, suppression, and combinations thereof — exposing more complex relationships between image and brain response than single-region maximization can reveal. Finally, in a prospective fMRI experiment with novel participants, NeuroVolve-generated stimuli targeting modulation of the amygdala drove neural responses consistent with predicted activation patterns,  demonstrating that our \emph{in silico} discoveries do generalize to \emph{in vivo} fMRI experimental findings, even in brain regions that are not primarily sensory.

\subsection*{NeuroVolve recovers known regional selectivity in visual cortex}

\begin{figure*}[!tp]
    \centering
    \includegraphics[width=\linewidth]{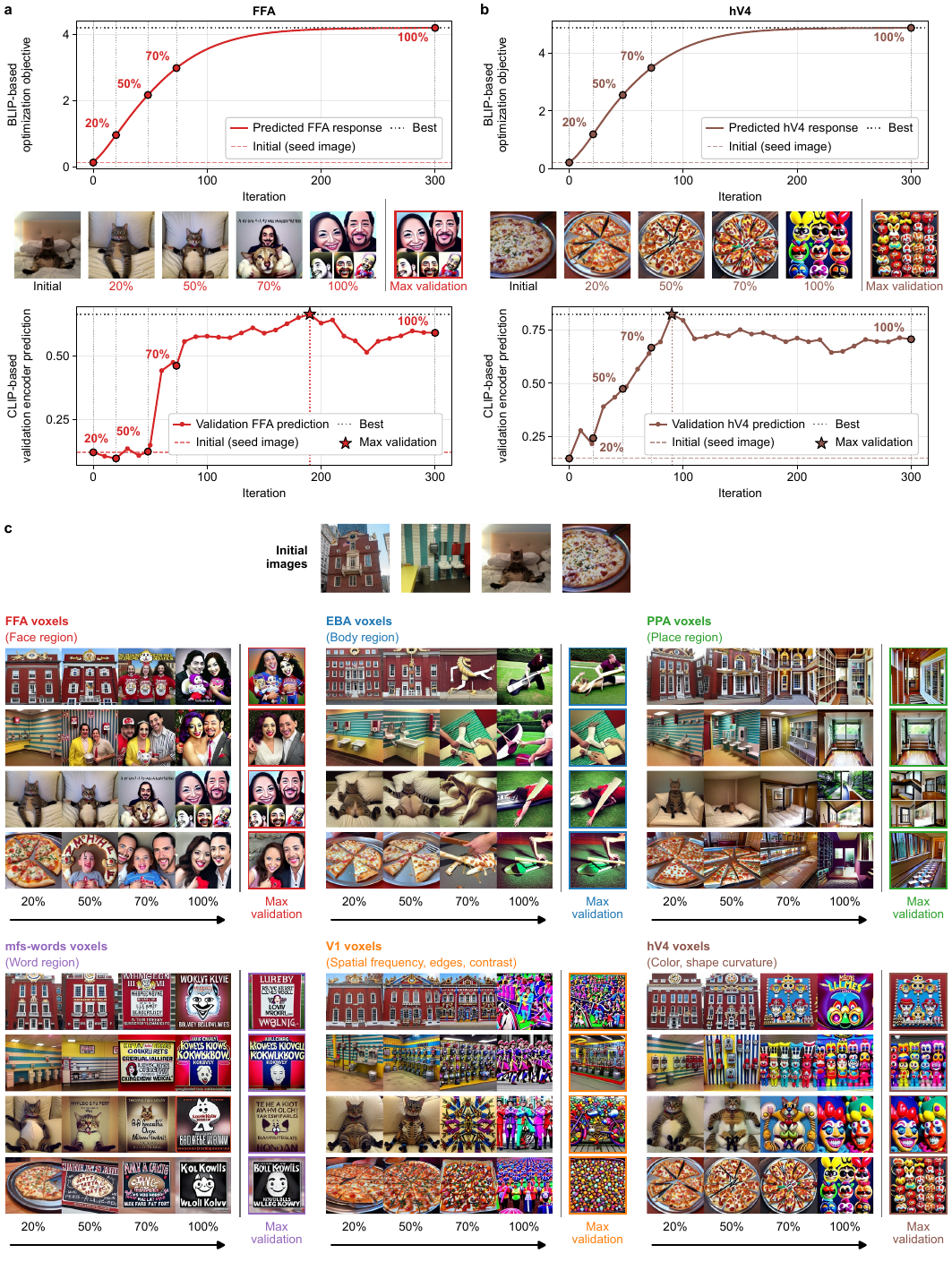}
    \caption{\textbf{NeuroVolve validation via a single-region maximization target objective.}
\textbf{a}, Optimization and validation trajectories for fusiform face area (FFA) from the initial cat seed image and \textbf{b} human visual area V4 (hV4) with the initial pizza seed image. Upper plots in both panels show the BLIP-based optimization objective value; lower plots show predictions of the subject-specific CLIP-based validation encoder, not used during optimization, for images generated along the trajectory. Points correspond to 20\%, 50\%, 70\% and 100\% of the total increase in the optimization objective, and the star marks the maximum value on the validation curve; the checkpoint images are shown between the plots.
\textbf{c}, Image sequences generated by maximizing the mean predicted voxel response within six visual regions of interest (ROIs): FFA, extrastriate body area (EBA), parahippocampal place area (PPA), mid-fusiform sulcus word-selective region (mfs-words), primary visual cortex (V1) and hV4. Each row has a different initial image (4 initial images shown at the top) and each column shows the 20\%, 50\%, 70\% and 100\% checkpoint images. The last column shows the generated image with the highest prediction from a separate CLIP-based validation encoder (max validation). Images are shown for NSD subject 5; other subjects are in the Supplementary Note~2.}
\label{fig:single_roi}
\end{figure*}

We first asked whether NeuroVolve could recover the established neural selectivity of individual brain regions using a simple objective: maximizing the mean predicted voxel response within a single target region. We examined six visual ROIs spanning the visual hierarchy: four higher-order category-selective regions---fusiform face area (FFA), extrastriate body area (EBA), parahippocampal place area (PPA), and mid-fusiform sulcus word area (mfs-words)---and two early visual regions (V1, hV4). For each ROI, we initialized optimization from the same four seed images and selected embeddings at iterations corresponding to 20\%, 50\%, 70\%, and 100\% of the total increase in the optimization objective and decoded them into images. 

Fig.~\ref{fig:single_roi}a and b shows representative optimization trajectories for two example regions FFA and hV4. In both examples, the target-ROI response predicted directly from the optimized Q-Former embeddings increased steadily before approaching a plateau. Images decoded along these trajectories showed progressive changes in visual content as the optimization objective increased. Because the optimization objective was evaluated directly in BLIP embedding space, an increase in this objective alone did not establish that the decoded images retained the predicted response increase. We therefore evaluated images generated along each trajectory using a separate subject specific CLIP-based validation encoder (for subject 5) that was not used to guide optimization (Methods). For both representative examples, the validation encoder's predicted responses mirrored the optimization curve (Fig.~\ref{fig:single_roi} a;b). Unlike the optimization objective, these validation trajectories were not constrained to increase monotonically, however their overall increase indicates that the targeted response changes predicted by the BLIP encoder optimization were preserved after image generation. Fig.~\ref{fig:single_roi}c shows the resulting image optimization sequence for all targeting regions for NSD subject 5. Other subjects' result are in the Supplementary Note~2

In all regions, the generated images increasingly exhibit features aligned with the known preferences of that target region. FFA-optimized images progressively emphasize faces\cite{kanwisher1997fusiform}, PPA converges on architectural and natural scene structures\cite{epstein1998cortical}, EBA generates body-part-like content\cite{downing2001cortical}, and mfs-words produces text-like patterns and letter forms\cite{cohen2000visual}. NeuroVolve also recovers visual features associated with earlier stages of visual processing. V1-optimized images exhibit high-frequency textures, edge-like structures, and high-contrast patterns\cite{deangelis1995receptive, devalois1990spatial}, while hV4-optimized images feature vivid colors, curvilinear shapes, and repetitive visual motifs, consistent with its reported sensitivity to color, complex form, and visual regularity.\cite{nandy2013fine, ziemba2016selectivity,kohler2016representation}.

\begin{figure*}[!htb]
    \centering
    \begin{subfigure}[t]{\linewidth}
        \includegraphics[width=\linewidth]{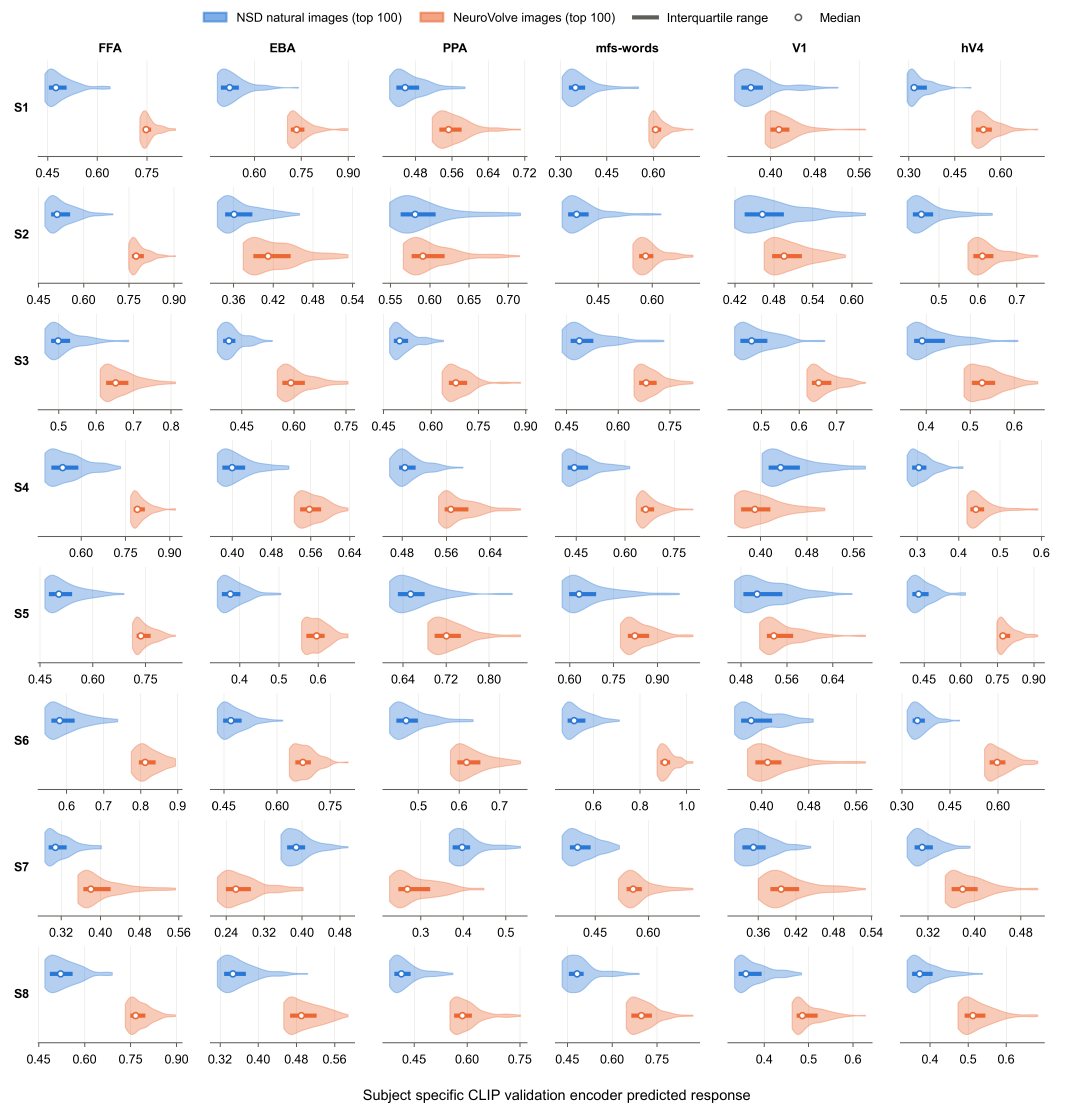}
    \end{subfigure}
    \caption{\textbf{CLIP-based validation-encoder's predicted responses to 100 top-ranked natural and NeuroVolve images.}
    Distributions show responses predicted by the participant-specific CLIP-based
    validation encoder for the 100 highest-ranked natural NSD images and 100
    highest-ranked NeuroVolve images in FFA, EBA, PPA, mfs-words, V1 and hV4.
    Circles indicate medians and thick bars indicate interquartile ranges.}
    \label{fig:violin_all}
\end{figure*}

We further quantified these observations using two large-scale analyses. First, zero-shot category classification showed that NeuroVolve images were more strongly aligned with their target semantic categories than the highest-ranked natural NSD images (with the ranking based either on the the fMRI-based measured responses or  encoding model predictions or the subject specific validation encoder) for faces, places and words, with comparable alignment for bodies (Supplementary Table~\ref{table:categ-clip}). Second, NeuroVolve images received higher encoder-predicted responses from the validation encoder than the highest-ranked natural images for most participants and regions (Supplementary Fig.~\ref{fig:violin_all}). Together, these analyzes indicate that NeuroVolve recovers established patterns of regional selectivity and identifies predicted maximum evoked-response stimuli beyond those represented in the natural-image dataset, without requiring the relevant visual features to be specified in advance.

\subsection*{NeuroVolve preserves within- and across-individual variability in maximally stimulating images}

\begin{figure*}[!tp]
    \centering
    \includegraphics[width=\linewidth]{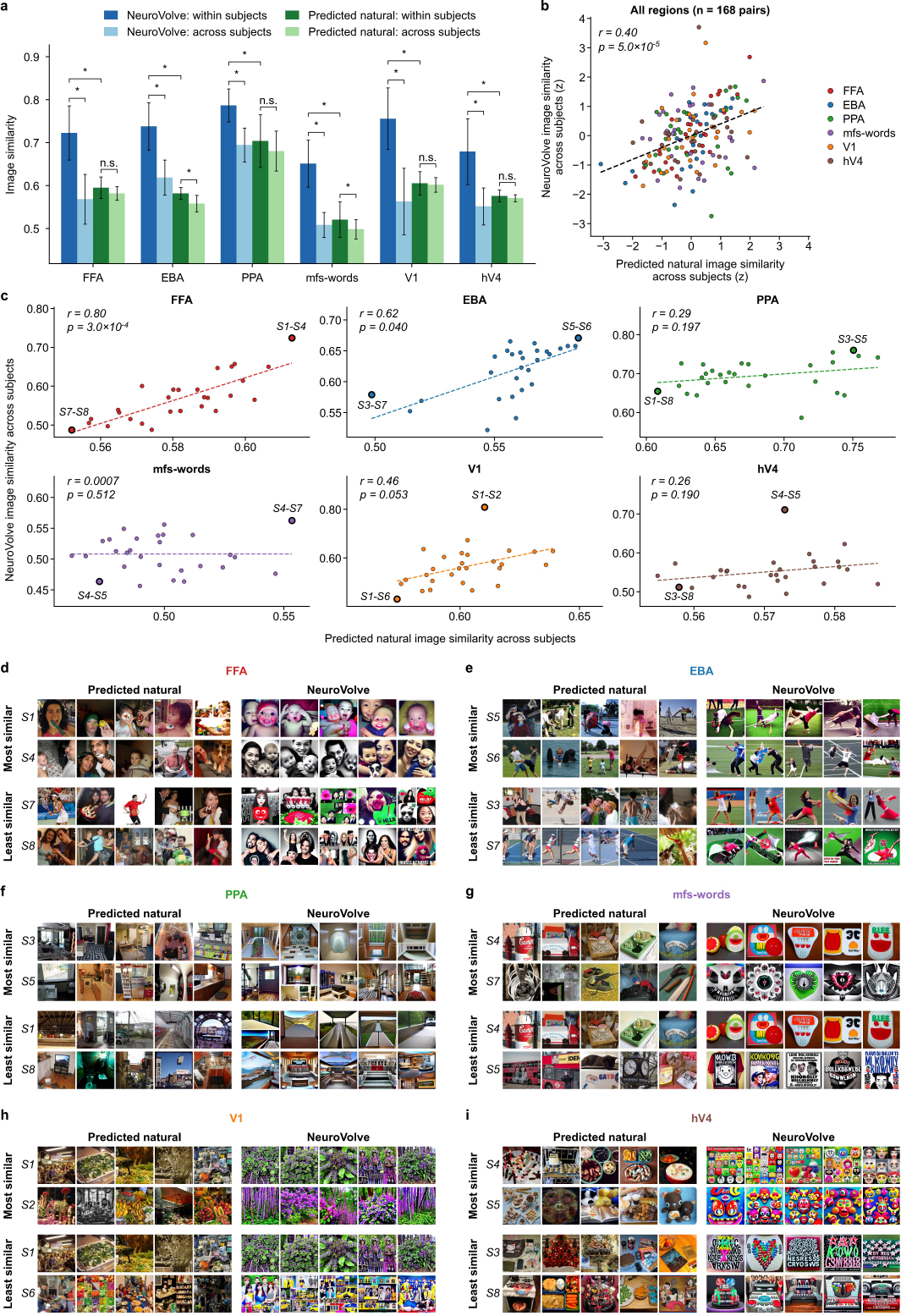}
\caption{\textbf{NeuroVolve preserves within- and across-individual differences in optimal images.}
\textbf{a}, Within- and across-subject similarity of the top 50 NeuroVolve images (blue) and the top 50 predicted natural NSD images (green) for each target region, ranked by the region's predicted activity from each subject's CLIP-based validation encoder. Similarity is the mean cosine similarity between image embeddings over all image pairs within or across subjects. Bars show the mean and error bars the s.d. across participants ($n=8$) or participant pairs ($n=28$). Brackets show two-tailed exact Wilcoxon signed-rank tests ($n=8$), with  values corrected across the 18 tests in \textbf{a} by the Benjamini--Hochberg false discovery rate procedure; *$q<0.05$; n.s., not significant.
\textbf{b}, Across-subject similarity of NeuroVolve versus natural images for all subject pairs and regions ($n=168$; 28 pairs $\times$ 6 regions), $z$-scored within each region. The dashed line shows the least-squares fit, and $r$ is the pooled Pearson correlation.
\textbf{c}, The same data as in \textbf{b}, shown for each region on the original similarity scale ($n=28$ pairs per panel). Outlined points mark the pairs shown in \textbf{d--i}. In \textbf{b} and \textbf{c}, $p$ values are from one-tailed exact permutation tests of participant labels.
\textbf{d--i}, The top five natural and NeuroVolve images of each participant in the most and least similar pairs in \textbf{c}, for FFA (\textbf{d}), EBA (\textbf{e}), PPA (\textbf{f}), mfs-words (\textbf{g}), V1 (\textbf{h}) and hV4 (\textbf{i}).}

\label{fig:subject_quant}
\end{figure*}

We next asked whether NeuroVolve could capture individual differences in maximally stimulating images beyond the broad stimulus categories associated with each region.
For each subject and region, we initialized optimization from the same 1,000 images drawn from the NSD "shared 1K" stimulus set included in every subject's image set.
Thus, all optimizations began from identical image set.Each trajectory was optimized for 300 steps, by which point the responses predicted by the BLIP-based optimization encoder had plateaued. The final embedding from each trajectory was then decoded to generate an image, yielding 1,000 NeuroVolved images per subject and region (Methods). As a natural-image reference, we used the 9,000 NSD images unique to each subject, excluding the shared 1,000 images so that exact image overlap would not affect across-subject image similarity measures.

For each subject, we scored the 1,000 NeuroVolved images and 9,000 subject-unique natural images using subject-specific CLIP-based validation brain encoders, different from the ones used during NeuroVolve optimization (see Methods). This allows for a separate brain activation-ranking model that is not related to NeuroVolve's image generation process. We retained the 50 images with the highest predicted regional responses from each pool. Thus, the natural and NeuroVolve sets were selected using the same proxy for each subject’s neural preference, while being drawn from independently obtained image pools. All subsequent analyses used these two top-50 sets for each region.

\begin{figure*}[!htb]
    \centering
    \begin{subfigure}[t]{\linewidth}
        \includegraphics[width=\linewidth]{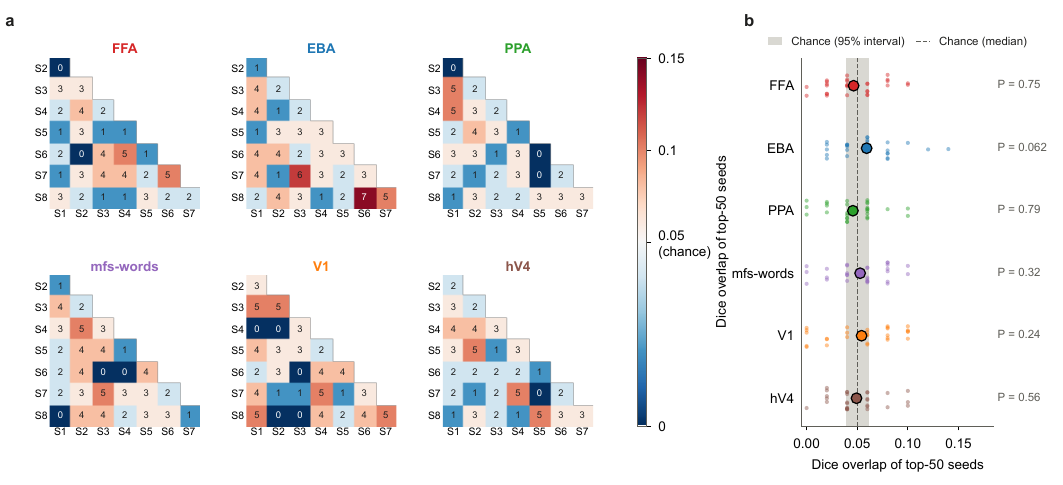}
    \end{subfigure}
   \caption{\textbf{Participants' top 50 NeuroVolve images do not share initial seed images beyond chance.}
\textbf{a}, Number of seeds shared by the top 50 NeuroVolve images of each participant pair, for each region; white marks the chance level of 2.5. \textbf{b}, Shared seeds for each pair (small dots; $n=28$) and their mean (large dots). The gray band and dashed line show the 95\% range and mean expected by chance (100,000 simulations of random top-50 selection). In no region was the overlap greater than chance (one-tailed uncorrected $p > 0.05$), so the fact that NeuroVolve images shared seeds had minimal effect on the subject-specificity results in Fig.~\ref{fig:subject_quant}.}
    \label{fig:overlap}
\end{figure*}

We first tested whether the selected images in the two sets were more similar within a subject than across subjects. We quantified image similarity as the cosine similarity between image embeddings (Fig.~\ref{fig:subject_quant}a; Methods). For the top 50 natural images, within-subject similarity was numerically higher than across-subject similarity in most regions, but the difference was statistically significant only in EBA and mfs-words. In contrast, the top 50 NeuroVolve images showed significantly higher within-subject than across-subject similarity in all six regions (all Benjamini-Hochberg corrected $p<0.05$). Within-subject similarity was also significantly higher for NeuroVolve images than for the selected natural images in every region (all corrected $p<0.05$). Importantly, the "shared 1K" seed images giving rise to each person's NeuroVolve top-50 sets were not overlapping across individuals at a rate higher than random chance (Supplementary Fig.~\ref{fig:overlap}). Thus, despite NeuroVolve's common initial seed image pool, the top 50 NeuroVolve images originated largely from different seeds across participants, suggesting that the shared initialization was unlikely effecting the across-subject image similarity metrics. It is worth noting that the final optimized images within a subject’s NeuroVolve set also all originated from different "shared 1K" seed images.
These results indicate that NeuroVolve optimization, even from different seed images, concentrated the final optimized images around a consistent set of visual features that appeared to be unique to each subject.

Greater consistency of generated images does not by itself establish that NeuroVolve reflects or amplifies across-subject differences, as it could be the case that the NeuroVolve images reflect convergence to similar sets of optimized images across subjects. We therefore next tested whether the across-subject relationships found in the top-50 natural image sets were preserved in the NeuroVolve image sets. That is, we tested the hypothesis that if two individuals had more similar top 50 natural images they also had more similar top 50 NeuroVolve images. For each region, we correlated the across-subject similarity matrix for the top 50 natural images with the corresponding matrix for the top 50 NeuroVolve images (Fig.~\ref{fig:subject_quant}b). Across-subject similarity was moderately correlated for the two image sources in a combined analysis of all brain regions together ($r=0.40, p = 5.0 \times 10^{-5}$, after z-scoring within each region). Within single regions, the across subject similarity varied, with higher values in FFA (Pearson \(r=0.80, p=3.0\times10^{-4}\)), EBA (\(r=0.62, p = 0.04\)), and V1 (\(r=0.46, p = 0.053\)) and weaker associations in PPA, mfs-words, and hV4 (all $p>0.05$)(Fig.~\ref{fig:subject_quant}c). For mfs-words, this lack of association could reflect a limit of the natural image set: few NSD (COCO) images contain only prominent text. In PPA, a possible reason is that scene images in COCO are relatively homogeneous, dominated by indoor settings such as kitchens and bathrooms, so subjects' top natural images vary little in ways that reflect individual scene preferences. Consistent with this, both the top 50 natural and top 50 NeuroVolve images for PPA had the highest across-subject similarity measures out of all the brain regions. In hV4, a possible reason is lower encoding accuracy in this smaller region, which could give a noisier optimization target and weaker discernible inter-individual differences. 

Thus, NeuroVolve did not simply generate arbitrary differences between subjects or converge on optimal stimuli that were similar for every subject. Instead, it preserved the across-subject structure inferred from top stimulating natural images, particularly in face- and body-selective regions. This organization is illustrated in the examples provided in (Fig.~\ref{fig:subject_quant}d–i), which show the top 5 NeuroVolve and top 5 natural images for individuals with high and low similarity of top stimulating images. For FFA, subjects 1 and 4 both had top natural images that contained young children - a characteristic that seemed to be accentuated in the NeuroVolve top images. On the other hand, the less-similar subjects 7 and 8 differed, with subject 7’s images containing adult faces with bright graphic and text-like elements and subject 8’s images predominantly showing interacting groups of adults. The EBA image sets for subjects 5 and 6 both emphasized bodies in motion on grassy sports fields, while subjects 3 and 7 had more dominant non-grass based sports and tennis, respectively. The PPA appeared to reflected differences in preferences for indoor/confined outdoor (subjects 3, 5 and 8) and outdoor transportation scenes (subject 1). Subject 4's mfs-words regions had top images of what appeared to be mostly cake/food decorated with text, and was most similar to subject 7 (tools) and least similar to subject 5 (objects with text, like signs and books, as expected for a word area). Differences also extended to lower-level visual properties. In V1, subjects 1 and 2 showed high-spatial-frequency, sketch-like patterns dominated by purple vegetation and plant-like forms, whereas subject 6’s images contained crowded groups of people with dominantly yellow and blue colors. In hV4, the most similar subjects shared curved shapes and repetitive visual structures while the least similar subjects appear to contain repeated letters/text and symmetric colorful forms. To assess whether these findings depended on using validation-encoder predictions to rank the natural images, we repeated the analysis using natural images ranked directly by measured fMRI responses. Despite the greater noise in measured-response-based rankings, the subject-pair similarity structure of the NeuroVolve and natural-image sets was significantly associated in FFA, with same-direction but nonsignificant associations in V1 and across regions. Full results are reported in Supplementary Note~\ref{note:subject-specificity} and Supplementary Fig.~\ref{fig:meas_subject}.

Together, these results support our argument that NeuroVolve converges on subject-specific top stimulating images that reflect measured across-subject similarity patterns in top stimulating natural images for some regions. The preservation of across-subject structure was clearest in FFA, EBA, and V1, indicating that NeuroVolve's ability to capture this across-subject specificity in top images varies depending on the visual region in question.

\subsection*{NeuroVolve for multi-region brain-response objectives}

\begin{figure*}[!tp]
    \centering
    \includegraphics[width=\linewidth]{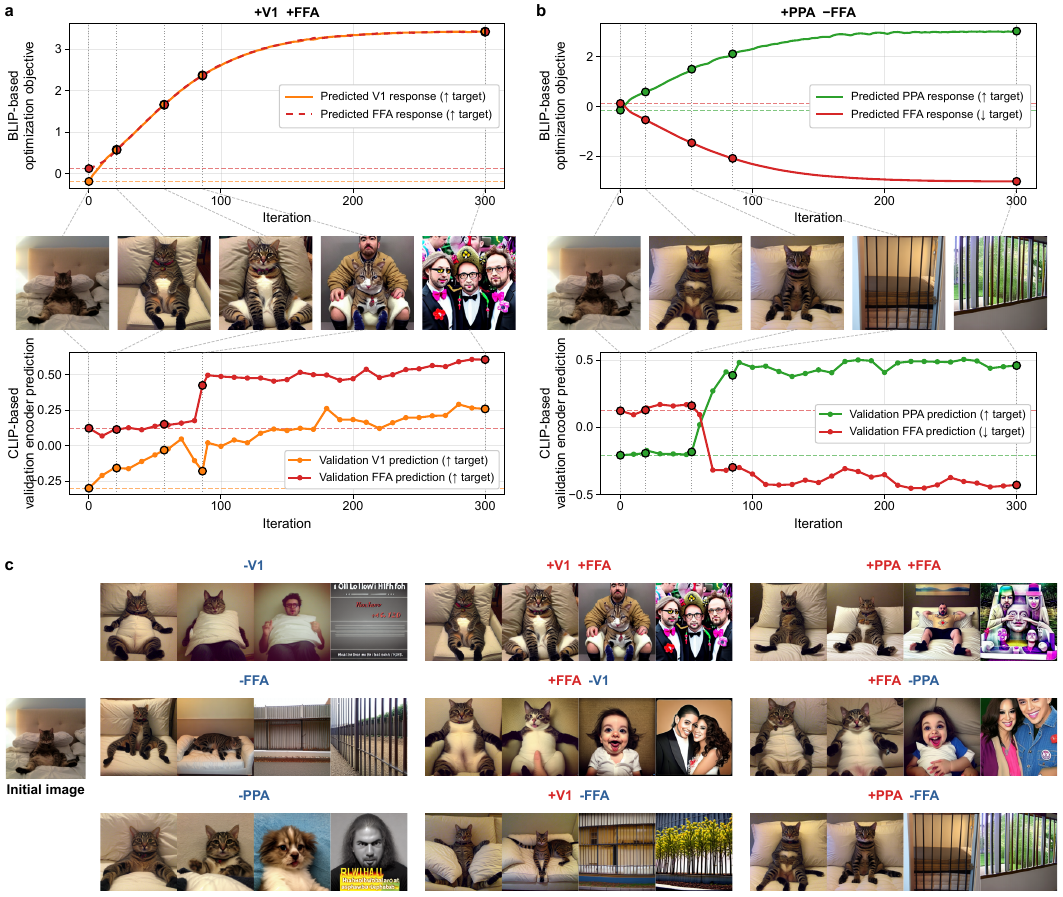}
   \caption{\textbf{Image synthesis under single- and multi-region brain-response objectives.}
\textbf{a,b}, Example trajectories for maximal co-activation of V1 and FFA (\textbf{a}) and maximal activation of PPA with minimal activation of FFA (\textbf{b}) starting from the same initial seed image (cat, on far left in row of images). \
Upper plots show the predicted response of each target region at every iteration of NeuroVolve's Q-Former embedding optimization, demonstrating optimization of the specific provided objective.
Lower plots show regional responses predicted by subject-specific CLIP-based validation encoder, not used during optimization, for images generated along each trajectory. 
Colored points mark selected checkpoints (initial value, 20, 50, 70 and 100$\%$ of final optimization curve value), with the checkpoints' corresponding generated images shown between the plots.  Dashed horizontal lines correspond to response to that regions' predicted response to the initial seed image. V1, primary visual cortex; FFA, fusiform face area; PPA, parahippocampal place area. Plus and minus signs denote maximization and minimization of the predicted regional response, respectively. \textbf{c}, NeuroVolve images generated from a single common seed image (cat, far left) under single-region minimization ($-$V1, $-$FFA, and $-$PPA), two-region co-maximization ($+$V1$+$FFA and $+$PPA$+$FFA), and mixed max/min two-region objectives ($-$V1$+$FFA, $+$V1$-$FFA, $-$PPA$+$FFA and $+$PPA$-$FFA). }
    \vspace{-0.1cm}
    \label{fig:multi_region}
\end{figure*}

We next demonstrated the flexibility of NeuroVolve to generate images for more complex, multi-region brain-response objectives. We considered three classes of objectives: minimizing a single target region, co-maximization of two regions, and maximizing one region while minimizing the other. These objectives were defined by either summing (co-max) or subtracting (mixed max/min) the mean predicted voxel responses of the target regions (Methods). We applied these objectives to three visual regions (V1, FFA, and PPA) resulting in 9 different multi-region objectives: three single-region minimization objectives and six two-region composite objectives. Fig.~\ref{fig:multi_region}a and b show two example optimization curves for our multi-region objective that show good behavior (V1 and FFA increasing in Fig.~\ref{fig:multi_region}a and PPA increasing and FFA decreasing in Fig.~\ref{fig:multi_region}b). Importantly, the subject-specific brain encoding validation model mirrored NeuroVolve's curves of expected max/min regional activations. Thus, the individual region responses changed in their intended directions according to a neural encoding model that was not used during optimization giving us confidence in our generated images achieving the intended neural objective.

Fig.~\ref{fig:multi_region}c shows the checkpoints for all 9 neural objectives from the same initial cat image. The generated images reflected anticipated combinations of known regional preferences. Co-activation of FFA and PPA produced images combining human faces with architectural or spatial context, whereas maximizing FFA while minimizing PPA produced images dominated by human faces with absent background content (plain background). Co-activation of V1 and FFA yielded images combining a multitude of human faces with high-frequency, colorful visual structure while maximizing FFA and minimizing V1 resulted in fewer faces with flatter texture and less color. The single-region minimization produced more heterogeneous images that lacked prominent features associated with the minimized region's preferences and also tended to be less colorful. This heterogeneity may reflect the broader range of images capable of producing low predicted responses compared with the narrower set that maximally activates a region.

Together, these results demonstrate NeuroVolve's ability to optimize complex brain-response objectives beyond single-region maximization. By flexibly combining activation and suppression targets, NeuroVolve enables the exploration of multi-dimensional neural response patterns \cite{gu2022neurogen, bashivan2019neural, gifford2025silico}, providing a framework for investigating how visual stimuli may differentially engage multiple brain regions within an individual.

\subsection*{Amygdala modulation with natural and NeuroVolve images}

\begin{figure*}[!tp]
\centering
        \includegraphics[width=\linewidth]{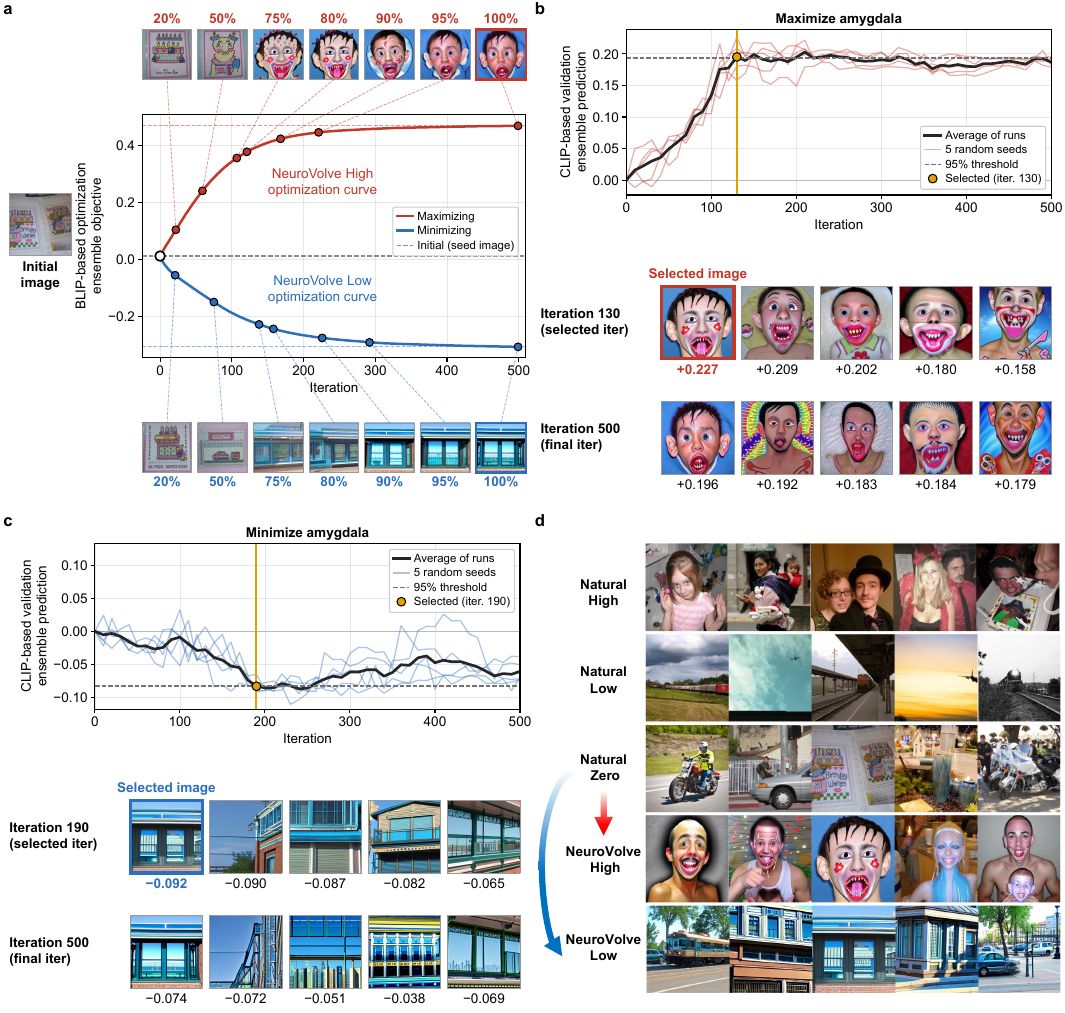}
    \caption{\textbf{Construction and \emph{in silico} validation of amygdala modulating images.} \textbf{a}, NeuroVolve optimization curve example of maximizing (red) and minimizing (blue) amygdala responses. Rows of images at top and bottom show corresponding images decoded at checkpoints that are the indicated percentages of the total change in the optimization objective. \textbf{b,c}, Validation model's predicted amygdala responses for five images generated with the same latent embedding but different guided diffusion seeds at each sampled iteration of the maximization (\textbf{b}) and minimization (\textbf{c}) trajectories. Thinner, colored lines show individual diffusion samples and thicker black lines their average. Horizontal dashed lines denote 95\% of the maximum magnitude predicted response change from baseline; yellow markers and vertical lines identify the earliest iteration reaching this 95\% threshold. Images at the bottom show samples from the early stopping selected iteration (top row) and final (bottom row) NeuroVolve iterations, with their validation ensemble predictions listed below them. Colored outlines identify an example image that was selected for the fMRI experiment. \textbf{d}, Five randomly selected images from: (top row) the top 65 natural images yielding the highest amygdala activity from the encoding model (Natural High), (second row) the top 65 natural images yielding the lowest predicted amygdala activity (Natural Low), and (third row) 65 natural images yielding predicted amygdala activity nearest to zero (Natural Zero). NeuroVolve High/Low images (rows 4 and 5) were initialized from Natural Zero images and optimized to increase or decrease the population-predicted amygdala response; images shown were selected based on their values in the CLIP-based validation population model prediction (see panels b, c). }
    \label{fig:amygdala_stimuli}
\end{figure*}

\begin{figure*}[!tp]
    \centering
    \includegraphics[width=\linewidth]{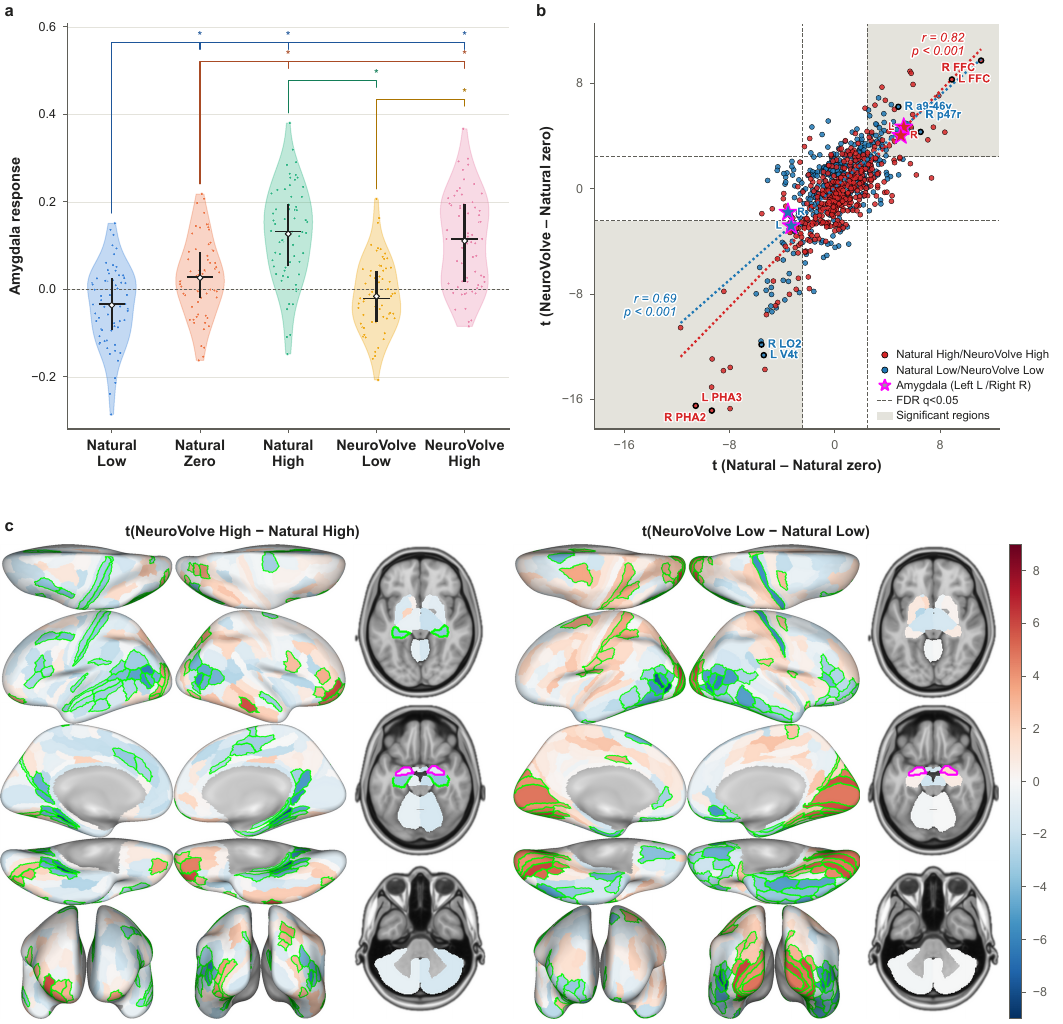}
   \caption{\textbf{\emph{in vivo} experimental validation of amygdala modulating images.}
Responses are shown as beta weights averaged across the amygdala and across six participants for a given image ($n=65$ images per condition, with two repeats per image).
\textbf{a}, Average amygdala responses per image, across five image conditions. Points represent image averages across participants; vertical bars are interquartile ranges; horizontal lines are medians; white diamonds are means; dashed line shows $\beta=0$. Brackets show significant Tukey HSD comparisons following ANOVA, adjusted across ten comparisons: $^{*}$corrected $p<0.05$.
\textbf{b}, The $x$ and $y$ axes show $t$-statistics of Natural high/low images' brain responses and NeuroVolve high/low images' brain responses, contrasted against the responses to Natural Zero images for 378 brain regions (360 HCP--MMP1 cortical parcels and 18 FreeSurfer subcortical structures). Each brain region contributes a High-condition point (red) and a Low-condition point (blue). Stars identify the left and right amygdala. Dotted lines show least-squares fits; $r$ denotes Pearson correlation. Dashed lines mark significance thresholds; shaded areas indicate significantly high or low activity for the corresponding contrasts. Extreme regions are labeled.
\textbf{c}, Regional $t$-statistics of brain activity in response to: NeuroVolve High images versus Natural High images (left) and NeuroVolve Low versus Natural Low (right). Red and blue indicate higher and lower NeuroVolve responses compared to Natural responses within the condition. Green outlines identify significantly different regions after Benjamini--Hochberg FDR correction (corrected $p<0.05$); pink outlines identify the amygdala.}
\label{fig:amygdala_cortical}

\end{figure*}

We conducted a prospective fMRI experiment to test whether natural and/or NeuroVolve-generated images could modulate responses in the amygdala, a region not primarily involved in visual processing but central to various psychiatric conditions like anxiety and depression. Candidate image generation and selection were guided by two separate population-level encoding ensembles, each formed by averaging predictions from the 8 NSD subject-specific encoding models. Population averaging was intended to reduce subject-specific variability and increase the chances of generalization to new participants. NeuroVolve optimization was guided by an average of the 8 subject level BLIP-based encoding models, with early stopping as described above using the average of the 8 subject-level CLIP-based validation encoding models. The natural image ranking used the average of the 8 subject-level CLIP-based validation models (Methods). 

Five sets of 65 images corresponding to five conditions were created: \emph{Natural High} (top 65 natural images by predicted activation), \emph{Natural 
Low} (65 natural images with lowest predicted activation), \emph{Natural Zero} (65 natural images with predicted activation closest to 0), \emph{NeuroVolve High} (65 NeuroVolve images created by maximizing predicted amygdala activation), and \emph{NeuroVolve Low} 
(65 NeuroVolve images created by minimizing predicted amygdala activation). Natural images were selected from all 72,000 participant-specific NSD images assigned to the eight participants, excluding the 1,000 shared images. NeuroVolve High and NeuroVolve Low images were generated by optimizing the 65 Natural Zero images towards high or low predicted responses (Fig.~\ref{fig:amygdala_stimuli}). Each NeuroVolve optimization had 500 iterations, with latent embeddings being used to reconstruct images every 10 iterations. At each sampled iteration, we decoded five images using independent diffusion seeds and passed them into the ensemble validation encoder. We then selected the earliest sampled iteration at which the mean predicted response from the ensemble validation encoder across the five decoded images reached 95\% of the maximum magnitude predicted response change. From the five images generated at that iteration, we retained the image with the highest predicted response for NeuroVolve High or the lowest predicted response for NeuroVolve Low (Fig.~\ref{fig:amygdala_stimuli} b,c; Methods). 

This image selection procedure used the ensemble validation encoder as an \emph{in silico} proxy for expected fMRI responses, reducing reliance on predictions from the BLIP encoding model that is based on a latent space and not directly on the image itself. Selecting an early checkpoint with a near-maximal predicted response change increases stimulus diversity. Images earlier on in the optimization curve tended to retain more features of their seed images, whereas later images often converged toward very similar visual content. The NeuroVolve selection process therefore balanced achieving the predicted response objective with retaining more variability in the stimuli presented to participants. 
 
Visual inspection of NeuroVolve-generated stimuli revealed distinct stimulus characteristics associated with maximizing and minimizing predicted amygdala responses (Fig.~\ref{fig:amygdala_stimuli}d). NeuroVolve High images predominantly contained human faces with strong emotional expressions, including smiles and open mouths and visible teeth, consistent with established associations between faces, emotion and amygdala activation \cite{adolphs1994impaired, morris1996differential} as well as the Natural High images. NeuroVolve Low images primarily contained buildings, vehicles and other human-made outdoor structures, with no human content, similar to the Natural Low images that contained outdoor transportation scenes. These images recovered expected amygdala-associated activating features while also, for the first time, suggesting visual content that is associated with suppressing amygdala responses.

\textbf{NeuroVolve images modulate amygdala responses \emph{in vivo}.}
In a prospective fMRI experiment, six novel participants (3 female, aged 32.3$\pm$11.7 years) viewed all 325 images (5 conditions $\times$ 65 images per condition) with two repeated viewings per image. As our primary outcome, we evaluated bilateral amygdala responses, with statistical inference performed across images after averaging responses across participants (Methods). Amygdala responses differed across the five conditions (one-way ANOVA, $F_{4,320}=38.67$, $p=2.1\times10^{-26}$; Fig.~\ref{fig:amygdala_cortical}a). Pairwise comparisons used Tukey HSD-adjusted $p$ values. Among the natural images, Natural High/Low elicited higher/lower responses than Natural Zero ($p=5.1\times10^{-8}$ and $p=0.002$, respectively). Similarly, NeuroVolve High also elicited higher responses than Natural Zero ($p=7.6\times10^{-6}$) and NeuroVolve Low ($p=4.1\times10^{-12}$). NeuroVolve Low had a lower mean response than Natural Zero, although this difference was not significant ($p=0.088$), indicating clearer evidence for upward rather than downward modulation. No significant differences existed between NeuroVolve High and Natural High ($p=0.870$) or between NeuroVolve Low and Natural Low ($p=0.741$).

\subsubsection*{Natural and NeuroVolve images produce similar but not identical regional response patterns}
We next examined whether the rest of the brain responded differently to natural and NeuroVolve image sets. Across 378 cortical and subcortical regions, we compared each High and Low condition with Natural Zero and assessed whether regions modulated by high/low natural images changed similarly in response to high/low NeuroVolve images (Fig.~\ref{fig:amygdala_cortical}b). Regional responses to each image were averaged across participants, and regional contrasts were computed across images ($n=65$ per condition).

Regional $t$-statistics were strongly correlated between natural and NeuroVolve images for both High ($r=0.82$, $p=5.6 \times 10^{-92}$) and Low ($r=0.69$, $p=2.2 \times 10^{-64}$) conditions. Sixty-three regions were significantly different from 0 in both High conditions and 67 in both Low conditions (FDR $q<0.05$), see gray quadrants in Fig.~\ref{fig:amygdala_cortical}b. All jointly significant regions changed in the same direction for both image sources, with agreement extending to ventral temporal, lateral occipitotemporal and frontoparietal regions, indicating correspondence in wide swaths of brain regions across non-visual areas. These correlations were treated as descriptive because the contrasts shared the Natural Zero baseline and regions were not statistically independent.

\begin{figure*}[!htb]
    \centering
    \begin{subfigure}[t]{\linewidth}
        \includegraphics[width=\linewidth]{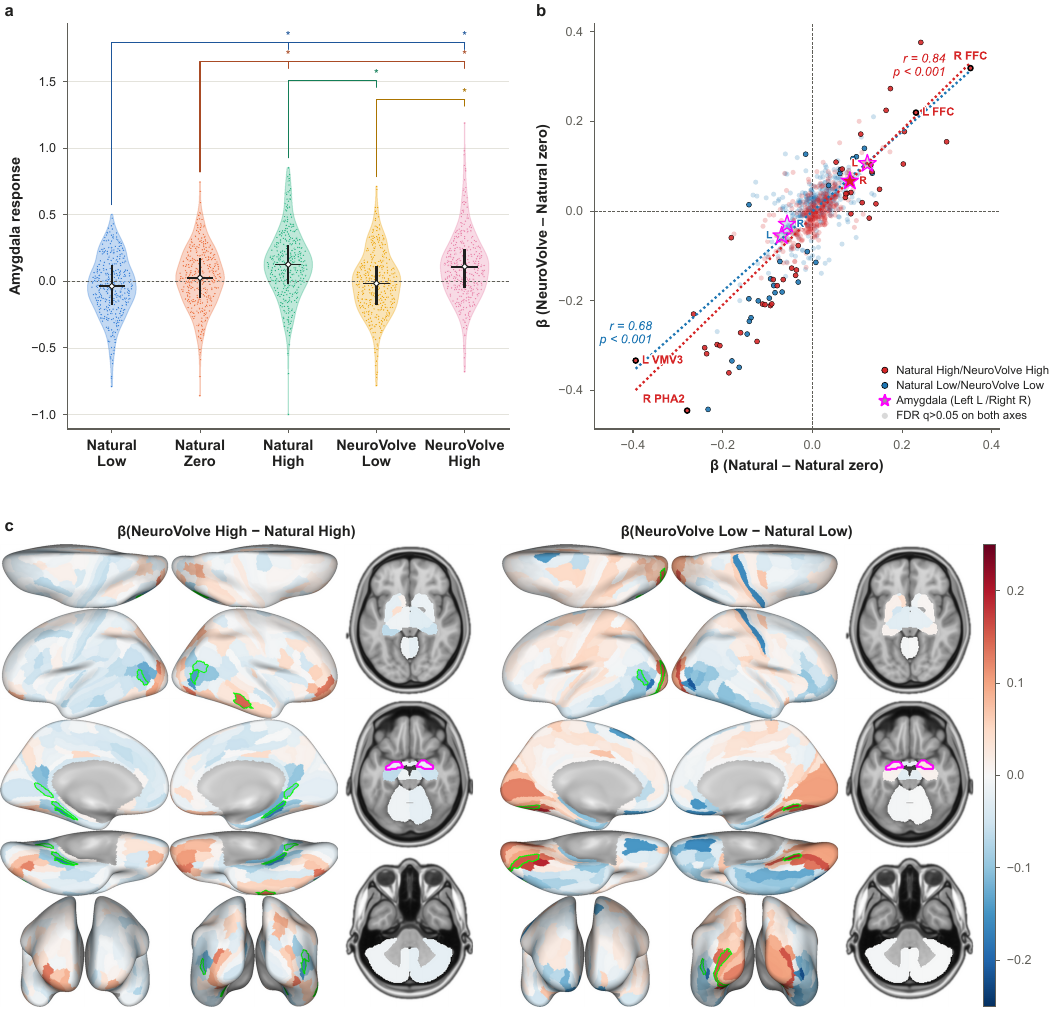}
    \end{subfigure}
    \caption{\textbf{Individual-level brain responses to amygdala modulating images.}
  Same layout as Fig.~\ref{fig:amygdala_cortical}, but without averaging image responses across participants. We used a linear mixed-effects model to test for image condition response differences, with participant and image as random effects. \textbf{a}, Amygdala responses for each image condition group, with points are individual participants' image responses for that condition. Vertical bars are interquartile ranges; horizontal lines are medians; white diamonds are means; dashed line shows $\beta=0$. Top brackets show significant pairwise contrasts from the mixed-effects model, Holm-adjusted across ten comparisons: $^{*}$corrected $p<0.05$. \textbf{b}, The $x$ and $y$ axes show mixed-effects model $\beta$ for Natural high/low images' brain responses and NeuroVolve high/low images' brain responses, contrasted against the responses to Natural Zero images for 378 brain regions, with the model fit separately for each region. Each brain region contributes a High-condition point (red) and a Low-condition point (blue). Solid points are significant for at least one of the two contrasts (Natural or NeuroVolve versus Natural Zero; Benjamini--Hochberg FDR-corrected $p<0.05$); faded points are significant for neither. Stars identify the left and right amygdala. Dotted lines show least-squares fits; $r$ denotes Pearson correlation. Extreme regions are labeled. \textbf{c}, Regional mixed-effects model $\beta$ of brain activity in response to: NeuroVolve High images versus Natural High images (left) and NeuroVolve Low versus Natural Low (right). Red and blue indicate higher and lower NeuroVolve responses compared to Natural responses within the condition. Green outlines identify significantly different regions after Benjamini--Hochberg FDR correction (corrected $p<0.05$); pink outlines the amygdala.}
    \label{fig:S_mixed}
\end{figure*}

Direct regional comparisons of Natural and NeuroVolve conditions identified differences in brain responses in 78 regions for High and 84 regions for Low (FDR $q<0.05$; Fig.~\ref{fig:amygdala_cortical}c), with differences in both directions. Some differences were shared across both comparisons: NeuroVolve images elicited higher responses than natural images in mid-level visual areas (V4, V8 and V3CD) for both High and Low conditions, with larger $t$-statistics for Low (for example, left V4: $t=2.97$ for High versus $t=6.05$ for Low; left V8: $t=3.85$ versus $t=5.10$; left V3CD: $t=3.85$ versus $t=4.38$; all FDR $q<0.05$). These shared effects may reflect visual properties of the generated images, such as their bright colors and textures. Natural images elicited higher responses in both conditions in lateral occipitotemporal cortex, specifically the MT+ complex (MT, MST and V4t; for example, left MST: $t=-5.48$ for High and $t=-7.76$ for Low), and the adjacent temporo-parieto-occipital junction (TPOJ2/3; FDR $q<0.05$). These regions are associated with processing visual motion, bodies and social interactions. The differences may reflect the visual content of the stimulus sets, including the greater presence of whole bodies, multiple people and ongoing actions in natural images. NeuroVolve High images elicited more suppressed brain responses than Natural High images in scene-selective brain regions, likely reflecting its emphasis on faces with reduced surrounding scene context. NeuroVolve High also elicited larger brain responses in object-selective cortex (PIT, LO1 and LO2), orbitofrontal and ventrolateral prefrontal cortex (bilateral 11l; right a47r, a10p, 10pp, IFSa and 47l), and right anterior temporal cortex (TE1m). Orbitofrontal cortex has been implicated in the top-down facilitation of object recognition \cite{bar2006top} this combination may reflect greater engagement of top down recognition processes when viewing NeuroVolve High images that are somewhat unnatural. NeuroVolve Low elicited higher responses than Natural Low in early visual cortex, which may reflect its prominent edges and textures. These results held when responses were pooled across participants and analysed with a mixed-effects model including participant identity (Supplementary Fig.~\ref{fig:S_mixed}).

Together, these findings indicate that NeuroVolve can take an image with near-average brain response (Natural Zero) and optimize it according to user-provided, target brain responses (e.g. amygdala up or down) to produce an image that actually elicits the target brain response in prospective fMRI experiments (e.g. amygdala up or down).

\section*{Discussion}
We introduced NeuroVolve, a data-driven generative framework for exploring 
participant-level neural selectivity and generating images that achieve user-defined target brain response patterns. This approach was validated by recovering established image feature selectivity across regions in the visual hierarchy and generating images that reflect and enhance measured across-subject variability in feature selectivity. Closing the empirical cycle, we validated NeuroVolve prospectively via fMRI experiments, demonstrating that our method's generated stimuli could even drive activity of a classically non-visual region (amygdala) in targeted ways.

Neural selectivity has traditionally been investigated by \emph{a priori} proposing visual features and designing stimuli to test them. NeuroVolve automates this hypothesis-generation step in a data-driven manner, loosening the restrictions of having to have \emph{a priori} knowledge. The embedding space of a pretrained vision-language model represents a space of candidate vision neuroscience hypotheses, while a neural encoding model scores candidates against a user-defined brain-response objective. Gradient-based optimization searches this space without requiring researchers to specify which visual features should satisfy the objective brain response pattern. This shifts the question from “does a predefined feature drive this region?” to “what visual content satisfies this brain response pattern?” This formulation builds on evidence that vision-language representations capture features predictive of human brain responses to natural scenes \cite{doerig2025nmi, conwell2024large}. NeuroVolve extends their use from response prediction to stimulus construction: optimized Q-Former embeddings are decoded into images using a frozen BLIP-Diffusion model \cite{li2023blip, li2023blipdiff}. Together, the encoding model and generator turn hypothesis generation into a tractable search over a broad, learned visual representation, producing candidate stimuli for subsequent experimentation that are not restricted by our prior knowledge and human-generated hypotheses.

Classic vision neuroscience has focused on identifying image features that maximally activate a given brain region of interest. In contrast, NeuroVolve can be flexibly used to generate images that achieve multi-region, multi-direction brain response patterns. Targeting co-activation of multiple regions allows identificaton of  content that supports multiple regional preferences, whereas opposing objectives seek content that increases one region's response while reducing another region's responses. Our validation experiments generated images that retained the targeted response directions using a separate brain encoding model not using within NeuroVolve, and resulted in images that reflected expected image feature combinations. Multi-region objectives offer a way to distinguish shared from region-specific visual preferences, including combinations difficult to sample from natural images because visual features often co-occur. Future experiments could include more complex targeted multi-region responses (more than 2 regions) or contain regional targets that are a specific value (rather than just maximizing or minimizing), allowing a much more unrestricted exploration of the brain's complex neural selectivity space.

Classically, vision neuroscience has focused on neural selectivity that is reproducible across a population of individuals. While of course this approach has merit, we are of the opinion that effort should also be made to map individual differences in neural selectivity \cite{gu2022neurogen, gu2022personalized, gu2023human}. Our previous work using NeuroGen, a precursor to NeuroVolve, showed that these types of tools can accentuate real, measured individual differences in visual feature preferences\cite{gu2022neurogen}, preserve inter-individual response relationships via encoding models trained on small data \cite{gu2022personalized}, and that individually tailored images can elicit stronger responses than population-based images or images tailored to other individuals in some higher-order visual regions\cite{gu2023human}. 

Building on these findings, we show here that NeuroVolve can be used to reveal individual differences in the features that drive brain regions' responses. We examined both the convergence of NeuroVolve to individually-consistent optimal stimuli features and preservation of across-subject similarity patterns of optimal stimuli features. NeuroVolve's maximally stimulating images derived from different initial image seeds were more similar than the maximally stimulating natural images, while their across-participant similarity structure corresponded to that of natural images most clearly in face- and body-selective cortex. These findings suggest that NeuroVolve's optimization can enhance individual visual preferences while preserving similarity relationships across participants, providing a viable route to personalized feature preference discovery. 

These results establish the framework's utility across regions and individuals within visual cortex, and provide motivation for the extension of these methods to brain regions whose visual preferences are less well characterized but whose activation patterns play an important role in emotional and mental health. Encoding-model-guided synthesis of optimal stimulating images has primarily targeted visual cortex\cite{bashivan2019neural,walker2019inception,ponce2019evolving,gu2022neurogen,ratan2021computational,luo2023brain,luo2023brainscuba,rose2024concentration,wang2022tuning}. Recent work has extended generative synthesis to computational models of the amygdala, characterizing the visual properties of stimuli optimized \emph{in silico}\cite{jang2025amygdala}. Our current work advances this direction by generating images for modulating the amygdala not just by maximizing activity but also minimizing it, and prospectively validating the images' predicted responses in \emph{in vivo} fMRI experiments. The additional prospective fMRI experiment also illustrates the complementary roles of \emph{in silico} exploration and \emph{in vivo} validation and testing. Encoding models and generators allow many candidate stimuli to be explored offline computationally before scanning\cite{naselaris2011encoding,gu2022personalized,iclrhmc}, while prospective fMRI experiments provide empirical validation of the candidate stimuli.  This workflow allows for more flexible, unrestricted, and efficient generation of complex stimuli designed for a given neural objective that in turn guide in a targeted way the more onerous and expensive fMRI experiments for empirical validation.  This combination could lower the cost of neuroscience discovery by enabling broad stimulus exploration and model-based validation before conducting new human experiments, concentrating experimental resources on testing the most promising computationally identified discoveries.

Several limitations of our work remain to be addressed in future work. NeuroVolve’s optimization is constrained by its pretrained visual representation, diffusion (image generation) model, and neural encoding models. NeuroVolve's optimization may therefore overlook visual properties that these models represent poorly or exploit prediction errors when handling out-of-sample images. As vision-language models continue to improve, we expect this ceiling to lift. The prospective fMRI experiment demonstrated that the NeuroVolve top images had amygdala responses that were higher than the Natural Zero images they were derived from, and the NeuroVolve Low images designed to suppress the amygdala, but they did not yield activation above and beyond the Natural High/Low images. Participant-specific models trained on larger datasets from a single individual could possibly support more reliable individualized modulation above and beyond a population level designed image set. Finally, the objectives examined here primarily averaged responses within regions. The voxelwise optimization encoder also permits objectives targeting individual voxels or distributed voxel patterns, offering a route to investigating selectivity differences obscured by regional averages. Such applications will require evaluation of voxel-level prediction reliability and extensive prospective validation.

In conclusion, NeuroVolve formulates optimal visual feature discovery as an inverse optimization problem, allowing researchers to predefine a targeted brain-response pattern without having knowledge of the visual features that should produce them. Subject-specific and multi-region objectives extend this search across individuals and more complex brain response combinations, while prospective validation in modulating the amygdala demonstrates this approach's application beyond visual cortex. By connecting data-driven stimulus generation with empirical testing, we demonstrate that NeuroVolve provides a flexible approach to discovering individual-level neural preferences beyond predefined stimulus categories and allows generation of candidate images outside of the restricted natural image sets used in large, publicly available databases.

\section*{Methods}
\subsection*{Dataset}
We used the Natural Scene dataset (NSD) \cite{allen2022massive} for much of our analysis, which includes whole-brain 7T fMRI data from 8 subjects who viewed 9000 - 10,000 naturalistic images from the MS COCO dataset with 2-3 repeats per image. A set of 1000 images were shared across all subjects (the so-called "shared 1K"), while the remaining images for each individual were mutually exclusive.

The fMRI data were pre-processed to correct for slice time differences and head motion using temporal interpolation and spatial interpolation. Then the single-trial beta weights representing the voxel-wise response to the image presented was estimated using a general linear model (GLM). There are three steps for the GLM: the first is to estimate the voxel-specific hemodynamic response functions; the second is to apply the GLMdenoise technique\cite{charest2018glmdenoise, kay2013glmdenoise}, to the single-trial GLM framework\cite{prince2022improving}; and the third is to use an efficient ridge regression\cite{rokem2020fractional} to regularize and improve the accuracy of the beta weights, which represent activation in response to the image. FreeSurfer was used to reconstruct the cortical surface, and both volume- and surface-based versions of the voxel-wise response maps were created. 

Data from the functional category localizer experiment (fLoc)\cite{stigliani2015temporal} was used to create contrast maps (voxel-wise t-statistics) of responses to specific object categories, and region boundaries were then manually drawn on inflated surface maps by identifying contiguous regions of high contrast in the expected cortical
location, and thresholding to include all vertices with contrast $>$ 0 within that boundary. Early visual ROIs were defined manually using retinotopic mapping data on the cortical surface. Surface defined ROIs were projected back to fill in voxels within the gray matter ribbon. Subcortical structures, specifically the individual-level amygdala masks were defined anatomically using FreeSurfer \cite{fischl2012freesurfer} automated subcortical segmentation (aseg)\cite{fischl2002whole} provided by NSD.

\subsection*{Encoding models}
NeuroVolve used two separately trained neural encoding models. The BLIP-based optimization encoder, denoted $\Phi_{\mathrm{opt}}$, predicted brain responses directly from Q-Former embeddings and provided the gradients used during embedding optimization within NeuroVolve. The CLIP-based validation encoder, denoted $\Phi_{\mathrm{val}}$, predicts responses directly from image CLIP embeddings and was used for post-optimization image valuation and selection; this model was totally independent from NeuroVolve's optimization process. 

\textbf{BLIP-based optimization encoder.} $\Phi_{\mathrm{opt}}$ used the frozen image encoder and Q-Former of BLIP-Diffusion\cite{li2023blipdiff}. The 16 Q-Former query embeddings (768 dimensions each) of an image were individually normalized to unit Euclidean norm and flattened into a 12,288-dimensional vector $q(I)$, which a multilayer perceptron (hidden layers of 2,048, 1,024 and 512 units, each followed by ReLU and dropout) and a linear readout mapped to voxelwise responses of retinotopic and category-selective visual ROIs:
\[
\widehat{\mathbf y}_{\mathrm{opt}}(I) = \Phi_{\mathrm{opt}}\!\left(q(I)\right).
\]
Only the multilayer perceptron and readout were trained.
\subparagraph*{Amygdala encoding model architecture.}
For the amygdala encoding model, $\Phi_{\mathrm{opt}}$ was replaced by a region-level variant, because voxel-level responses in the amygdala are noisy owing to its low signal-to-noise ratio. With the same frozen backbone and input normalization, a multilayer perceptron (hidden layers of 1,024, 512 and 256 units) and a nonlinear readout predicted the average response (beta weights) of the voxels in the left and right amygdala.

\textbf{CLIP-based validation encoder.}
$\Phi_{\mathrm{val}}$ used the frozen OpenCLIP ViT-H/14 image encoder trained on LAION-2B\cite{cherti2023openclip}. The 1,024-dimensional image embedding $f_{\mathrm{CLIP}}(I)$ was normalized to unit Euclidean norm and mapped by a linear readout to the voxel-averaged responses of the same retinotopic and category-selective visual ROIs, with one output per ROI:
\[
\widehat{\mathbf y}_{\mathrm{val}}(I)
    = W \frac{f_{\mathrm{CLIP}}(I)}{\lVert f_{\mathrm{CLIP}}(I)\rVert_2} + b.
\]
Only the readout was trained.

\subparagraph*{Amygdala validation encoding model.}
The amygdala's validation encoder $\Phi_{\mathrm{val}}$ uses the same frozen backbone and normalization and training procedure, but instead predicts the average of the voxel-wise beta weights (responses) in the left and right amygdala. 

\textbf{Model training and evaluation.}
Within each subject, responses to repeated presentations of the same image were averaged before model training. The NSD "shared 1K" images viewed by all subjects were used for final evaluation. The remaining subject-specific images (up to approximately 9,000, depending on the number of completed scan sessions) were used for model development. For each subject, 500 of these images were assigned to a validation set, and the remaining images constituted the training set.

The optimization encoder, $\Phi_{\mathrm{opt}}$, was trained separately for each subject by minimizing the mean-squared error between predicted and measured voxelwise or regional responses. We used AdamW with a learning rate of $3\times10^{-4}$, a batch size of 64, and gradient-norm clipping at 1.0. Because the 16 Q-Former query tokens are not associated with fixed spatial locations, their ordering has no inherent meaning. We therefore randomly permuted the tokens for each training sample, encouraging $\Phi_{\mathrm{opt}}$ to be approximately invariant to token order and preventing subsequent embedding optimization from exploiting order-specific encoder features. Prediction accuracy on the validation set was assessed after each epoch. Training was stopped if accuracy failed to improve for five consecutive epochs, and the checkpoint with the highest validation accuracy was retained.

The CLIP-based validation encoder, $\Phi_{\mathrm{val}}$, was trained separately for each subject using all available subject-specific images, including the 500 images used for validation of $\Phi_{\mathrm{opt}}$. No separate validation split or early stopping was used for $\Phi_{\mathrm{val}}$. Training ran for 100 epochs with a weight decay of 0.02 and an exponentially decaying learning rate from $3\times10^{-4}$ to $1.5\times10^{-4}$. The checkpoint from the final epoch was retained.

Final prediction accuracy was evaluated on the held-out "shared 1K" images and quantified as the Pearson correlation between predicted and measured responses across images. The modeled voxels and regions, together with the prediction accuracy of each encoder, are reported in Supplementary Note 1.

\subsection*{Q-Former embedding optimization}

NeuroVolve optimizes a Q-Former embedding to satisfy a user-defined brain-response objective. For a target region $R$, we defined its predicted response as the mean response predicted by the BLIP-based optimization encoder across its voxels:
\[
m_R(q)
=
\frac{1}{|R|}
\sum_{v\in R}\Phi_{\mathrm{opt}}(q)_v.
\]

\textbf{Brain-response objective formulations.} We considered four types of brain-response objectives:

\begin{itemize}

\item \textbf{Maximize the predicted response in region $i$:}
\[
\mathcal{L}_{\mathrm{brain}}(q)
=
-m_i(q).
\]

\item \textbf{Suppress the predicted response in region $i$:}
\[
\mathcal{L}_{\mathrm{brain}}(q)
=
m_i(q).
\]

\item \textbf{Co-activate regions $i$ and $j$:}
\[
\mathcal{L}_{\mathrm{brain}}(q)
=
-\left[
\lambda_1 \bigl(m_{i}(q) + m_{j}(q)\bigr)
-
\lambda_2 \bigl\lvert m_{i}(q) - m_{j}(q) \bigr\rvert
\right].
\]

\item \textbf{Activate region $i$ while suppressing region $j$:}
\[
\mathcal{L}_{\mathrm{brain}}(q)
=
-\left[
\lambda_1\bigl(m_{i}(q) - m_{j}(q)\bigr)
-
\lambda_2 \bigl\lvert m_{i}(q) + m_{j}(q) \bigr\rvert
\right].
\]

\end{itemize}

For multi-region objectives, the first term drives both regions in the desired directions, while the second discourages either response from dominating. Specifically, it promotes matched responses for co-activation and equal-magnitude, opposite-sign responses for activation–suppression. Here, \(m_i(q)\) denotes the mean predicted response across all voxels in region \(i\); subdivisions (e.g., FFA1/FFA2 and V1v/V1d) were pooled. We set \(\lambda_1=\lambda_2=0.5\) in all experiments.

\textbf{Optimization procedure.}
The initial embedding $q^0$ was extracted from an initialization image using the frozen BLIP image encoder and Q-Former. Only the Q-Former embedding was updated; all model parameters remained frozen. The selected brain-response objective loss was minimized using Adam with a learning rate of $0.01$.

\subsection*{NeuroVolve's image generation}

Optimized Q-Former embeddings were decoded using the frozen BLIP-Diffusion model \cite{li2023blipdiff}. The 16 Q-Former query tokens were inserted into the text-encoder input alongside an empty text prompt, and the resulting representation conditioned the frozen diffusion U-Net through cross-attention. Images were generated at $512\times512$ pixels using 50 denoising steps of the PNDM sampler with its default scaled-linear noise schedule and a classifier-free guidance scale of 7.5. The unconditional branch used an empty prompt without query tokens. For each of NeuroVolve's optimization trajectories, one diffusion seed was sampled and held fixed across all decoded embeddings so that changes along the trajectory primarily reflected embedding optimization rather than diffusion noise. In the prospective amygdala experiment, five independent diffusion seeds were used per trajectory and held fixed across sampled optimization iterations.

\subsection*{Subject-specific image analysis}

For each NSD participant and each target region (EBA, FFA, PPA, mfs-words, V1 and hV4), we generated 1,000 NeuroVolve images by maximizing the response predicted by that participant's BLIP-based optimization encoder $\Phi_{opt}$. The same 1,000 initialization images from the "shared 1K" image set were used for every participant so that differences between the resulting image sets could not be attributed to different initialization sets. Optimization was run for 300 steps, by which point the BLIP predicted-response trajectories had plateaued across participants and target regions, indicating that the optimization had converged. One image decoded from the final optimized embedding was retained from each trajectory. The 1,000 NeuroVolved images were ranked using the corresponding participant-specific CLIP based validation encoder $\Phi_{val}$, and the 50 images with the highest predicted responses were retained. As a natural-image set comparison, we ranked the 9,000 NSD images unique to that participant using the same validation encoder and retained the 50 highest-ranked images. 

Similarity between pairs of images was measured by taking the cosine similarity of each images' embeddings in the latent space of a frozen OpenAI CLIP ViT-B/32 image encoder\cite{radford2021clip}. For each participant, within-participant similarity was the mean across the $\binom{50}{2}=1,225$ distinct image pairs in that participant's set of top 50 images. For each of the $\binom{8}{2}=28$ participant pairs, across-participant similarity was the mean across the $50\times50=2,500$ image pairs within each subjects' top 50 image sets. This calculation was performed separately for natural and NeuroVolve images.

To test whether the across-subject similarity patterns found within the natural-image sets was preserved in the NeuroVolve image sets, we calculated Pearson's correlation between the upper triangles of the corresponding $8\times8$ similarity matrices. One-tailed $p$ values were calculated via an exact permutation test in which participant labels were permuted across all $8!=40{,}320$ orderings.

To summarize across-subject similarity across multiple brain regions, participant-pair similarities were $z$-scored within each region before
pooling the $6 \times 28 = 168$ values, so that the pooled correlation, which equals the mean of the
region-wise correlations, reflects which participant pairs were similar rather than differences in
similarity level between regions; the same subject-pair permutation was applied to all regions simultaneously. For
illustrative purposes, the most and least similar participant pairs in each region were those with the largest
and smallest sum of the $z$-scored natural and NeuroVolve similarities.

\subsection*{Prospective amygdala modulation experiment}
\textbf{Population-level encoding ensembles.} We constructed separate population-level ensembles for the BLIP-based optimization encoders $\Phi_{opt}$ and the CLIP-based validation encoders $\Phi_{val}$. For an image or Q-Former embedding, predictions were obtained from each of the eight participant-specific models and averaged across participants. The amygdala score was calculated as the mean of the predicted left and right amygdala responses. The BLIP-based ensemble $\Phi_{opt}$ was used exclusively to optimize NeuroVolve embeddings. The CLIP-based ensemble $\Phi_{val}$ was used to rank natural images and to evaluate and select from NeuroVolve's generated images. Thus, NeuroVolve's final image set is selected based on an encoding model prediction that was not used during optimization, and that is based on the reconstructed image itself rather than a latent image embedding. The use of the population-level encoding predictions provides a participant-independent estimate of amygdala response that had the best likelihood of generalizing to novel participants.

\textbf{Natural image set selection}
The candidate pool comprised all 72,000 participant-specific NSD images assigned to the eight participants (9,000 per participant), excluding the 1,000 shared images. Each image was scored using the population-level CLIP-based validation ensemble. From this set, we selected i) the top 65 natural images that maximized predicted amygdala response for the population-level ensemble encoder (called “Natural High"), ii) the 65 natural images that minimized predicted amygdala response for the population-level ensemble encoder (called “Natural Low") and iii) the 65 natural images that minimized the absolute value of predicted amygdala response, i.e., achieved an response that was as close to average (0) as possible, for the population-level ensemble encoder (called “Natural Zero"). 

\textbf{NeuroVolve image set selection}
Each of the 65 "Natural Zero" images was used to initialize NeuroVolve and optimize for high or low amygdala activation. The Q-Former embeddings were optimized for 500 iterations using the population-level BLIP-based ensemble and sampled every 10 iterations. At each sampled iteration, 5 images were generated using different random seeds for the diffusion model. Every decoded image was evaluated using the population-level validation ensemble, and the five predicted responses were averaged to obtain an image-space response validation trajectory. For each validation trajectory, we selected the earliest sampled iteration at which the mean validation prediction reached 95\% of the largest predicted response change observed along that trajectory. For maximization, this was the earliest iteration reaching 95\% of the increase from the initial prediction to the maximum prediction. For minimization, it was the earliest iteration reaching 95\% of the decrease from the initial prediction to the minimum prediction. Among the five images decoded at the selected iteration, the image with the highest/lowest validation prediction was retained in the "NeuroVolve High/Low" image sets. Repeating this procedure for each of the 65 Natural Zero seed images produced 65 "NeuroVolve High" and 65 "NeuroVolve Low" images.

\textbf{fMRI experimental design and analysis}
We collected prospective data from seven individuals (4 female, aged 31.7$\pm$10.7 years) using a GE Discovery MR750 3T MRI scanner, including anatomical T1 and whole-brain fMRI acquisitions. Functional MRI data consisted of 60 oblique axial slices with \(2.03 \times 2.03 \times 2.4\ \mathrm{mm}\) voxels, \(\mathrm{TR}/\mathrm{TE} = 1950/32\ \mathrm{ms}\), in-plane acceleration factor 2, multiband factor 2, and posterior-to-anterior (P\(\gg\)A) phase encoding. Subjects viewed a series of \(8.4^\circ\) square images, each presented for 3 seconds (2 s on, 1 s off), organized into blocks of 14 images within a condition. Subjects maintained fixation on a central dot and performed a 1-back task with button press to increase aletness and maintain attention. A single 254-second run consisted of five 42-second stimulus blocks separated by 6 seconds of rest. Each session consisted of 10 image-viewing fMRI scans. Anatomical scans (T1 MPRAGE, 1mm voxels) were processed using FreeSurfer 7.4.1. Preprocessing for fMRI consisted of 1) a combined slice-time correction and upsampling to 1.0 sec TR using a custom script 2) motion estimation using FSL MCFLIRT 3) customized fieldmapless distortion correction and coregistration to anatomy using FreeSurfer SynthMorph\cite{hoffmann2024anatomy}, and finally a one-step resampling combining all of these, resampling to an anatomically-aligned 2mm isotropic grid. Motion for each run was summarized using framewise displacement, and one subject was excluded from further analysis due to significant motion throughout the session (maximum framewise displacement per run averaged $>$ 2mm). For the remaining 6 individuals (3 female, aged 32.3$\pm$11.7 years), the response to each individual image was estimated from the preprocessed time series using the GLMsingle package. GLM estimates for one-back trials were excluded from analysis. Voxelwise GLM beta coefficients were averaged across voxels within a given region of interest and subsequently z-scored across all images within each subject. Each subject's cortex was divided into 360 regions using the HCP Multi-Modal Parcellation \cite{glasser2016multi}. The amygdala masks and other subcortical regions were defined anatomically using FreeSurfer automated subcortical segmentation (aseg)\cite{fischl2002whole}.

\textbf{Statistical information}
In Fig.~\ref{fig:subject_quant}a, two-tailed exact Wilcoxon matched-pairs signed-rank tests ($n = 8$ participants) compared within- with across-participant similarity, and NeuroVolve with natural within-participant similarity. For within-versus-across
comparisons, each participant's within-participant similarity was paired with their mean across-participant similarity over their seven pairings with the other participants. The resulting 18 $p$ values (6 regions $\times$ 3 comparisons) were corrected together with the Benjamini--Hochberg false discovery rate procedure \cite{benjamini1995controlling}. Error bars show the s.d. across the 8 participants (within-participant) or the 28 participant pairs (across-participant). In Fig.~\ref{fig:subject_quant}b,c, one-tailed $p$ values for Pearson correlations between natural- and NeuroVolve-image similarity across participant pairs ($n = 28$ per region; $n = 168$ pooled across six regions in b) were calculated with an exact permutation test of participant labels ($8! = 40{,}320$ permutations), applying the same permutation to all regions in b. In Fig.~\ref{fig:amygdala_cortical}a, a one-way ANOVA across the five image conditions ($F(4, 320) = 38.67$) was followed by two-tailed Tukey's HSD tests for all pairwise comparisons ($n = 65$ images per condition, responses averaged across 6 participants). In Fig.~\ref{fig:amygdala_cortical}b and c, two-tailed independent two-sample $t$-tests ($n = 65$ vs. 65 images, d.f. $= 128$) were computed for each of 378 regions, and Benjamini-Hochberg FDR correction across regions was used. Statistical differences were considered significant if $p < 0.05$ (FDR $q < 0.05$). All $p$ values reported in the paper are two-tailed unless otherwise stated.

\section*{Data availability}

The Natural Scenes Dataset (NSD) is publicly available at
\url{http://naturalscenesdataset.org}. The Neurovolve Dataset is available at \url{https://doi.org/10.5281/zenodo.23780000}.

\section*{Code availability}
NeuroVolve code is available at \url{https://github.com/haomiao8/NeuroVolve}.

\backmatter

\bmhead{Supplementary information}
Supplementary information is provided in a separate file.

\bmhead{Acknowledgments}
Funding was provided in part by Weill Cornell Medicine's Department of Radiology through a Cornell University MRI Facility Seed Grant. We further acknowledge the help of the Cornell MRI Facility in image acquisition.

\bmhead{Author contributions}
A.K., M.R.S., H.C. and K.W.J. conceived and conducted the experiments and interpreted the results. H.C. additionally analyzed the results and carried out the statistical analyses. K.W.J. additionally processed the imaging data and designed the MRI experimental protocols. K.W.J., A.K., D.C., Y.Z. and H.C. conducted data collection. H.C. and A.K. wrote the manuscript. All authors reviewed the manuscript.

\bmhead{Competing interests}
The authors declare no competing interests.

\bibliography{sn-bibliography.bib}

@article{doerig2025nmi,
  title   = {High-level visual representations in the human brain are aligned with large language models},
  author  = {Doerig, Adrien and Kietzmann, Tim C. and Allen, Emily and Wu, Yihan and Naselaris, Thomas and Kay, Kendrick and Charest, Ian},
  journal = {Nature Machine Intelligence},
  volume   = {7},
  number   = {8},
  pages    = {1220--1234},
  year     = {2025},
  publisher= {Nature Publishing Group},
  doi      = {10.1038/s42256-025-01072-0}
}

@article{conwell2024large,
  title={A large-scale examination of inductive biases shaping high-level visual representation in brains and machines},
  author={Conwell, Colin and Prince, Jacob S and Kay, Kendrick N and Alvarez, George A and Konkle, Talia},
  journal={Nature communications},
  volume={15},
  number={1},
  pages={9383},
  year={2024},
  publisher={Nature Publishing Group UK London}
}

@article{morris1996differential,
  title={A differential neural response in the human amygdala to fearful and happy facial expressions},
  author={Morris, J S and Frith, C D and Perrett, D I and Rowland, D and Young, A W and Calder, A J and Dolan, R J},
  journal={Nature},
  volume={383},
  number={6603},
  pages={812--815},
  year={1996}
}

@article{adolphs1994impaired,
  title={Impaired recognition of emotion in facial expressions following bilateral damage to the human amygdala},
  author={Adolphs, R and Tranel, D and Damasio, H and Damasio, A},
  journal={Nature},
  volume={372},
  number={6507},
  pages={669--672},
  year={1994}
}

@article{gu2022personalized,
  title={Personalized visual encoding model construction with small data},
  author={Gu, Zijin and Jamison, Keith and Sabuncu, Mert and Kuceyeski, Amy},
  journal={Communications Biology},
  volume={5},
  number={1},
  pages={1382},
  year={2022},
  publisher={Nature Publishing Group UK London}
}

@article{nandy2013fine,
  title={The fine structure of shape tuning in area V4},
  author={Nandy, Anirvan S and Sharpee, Tatyana O and Reynolds, John H and Mitchell, Jude F},
  journal={Neuron},
  volume={78},
  number={6},
  pages={1102--1115},
  year={2013},
  publisher={Elsevier}
}

@article{ziemba2016selectivity,
  title={Selectivity and tolerance for visual texture in macaque V2},
  author={Ziemba, Corey M and Freeman, Jeremy and Movshon, J Anthony and Simoncelli, Eero P},
  journal={Proceedings of the National Academy of Sciences},
  volume={113},
  number={22},
  pages={E3140--E3149},
  year={2016},
  publisher={National Academy of Sciences}
}

@article{devalois1990spatial,
  title={Spatial vision},
  author={DeValois, Russell L and DeValois, Karen K},
  year={1990},
  publisher={Oxford university press}
}

@article{kanwisher1997fusiform,
  title={The fusiform face area: a module in human extrastriate cortex specialized for face perception},
  author={Kanwisher, Nancy and McDermott, Josh and Chun, Marvin M},
  journal={Journal of neuroscience},
  volume={17},
  number={11},
  pages={4302--4311},
  year={1997},
  publisher={Society for Neuroscience}
}

@article{epstein1998cortical,
  title={A cortical representation of the local visual environment},
  author={Epstein, Russell and Kanwisher, Nancy},
  journal={Nature},
  volume={392},
  number={6676},
  pages={598-601},
  year={1998},
  publisher={Nature Publishing Group UK London}
}

@article{Ishai1999,
author = {Ishai, A and Ungerleider, L G and Martin, A and Schouten, J L and Haxby, J V},
journal = {Proc Natl Acad Sci U S A},
number = {16},
pages = {9379--9384},
title = {{Distributed representation of objects in the human ventral visual pathway.}},
volume = {96},
year = {1999}
}

@article{downing2001cortical,
  title={A cortical area selective for visual processing of the human body},
  author={Downing, Paul E and Jiang, Yuhong and Shuman, Miles and Kanwisher, Nancy},
  journal={Science},
  volume={293},
  number={5539},
  pages={2470--2473},
  year={2001},
  publisher={American Association for the Advancement of Science}
}

@article{naselaris2011encoding,
  title={Encoding and decoding in fMRI},
  author={Naselaris, Thomas and Kay, Kendrick N and Nishimoto, Shinji and Gallant, Jack L},
  journal={Neuroimage},
  volume={56},
  number={2},
  pages={400--410},
  year={2011},
  publisher={Elsevier}
}

@article{allen2022massive,
  title={A massive 7T fMRI dataset to bridge cognitive neuroscience and artificial intelligence},
  author={Allen, Emily J and St-Yves, Ghislain and Wu, Yihan and Breedlove, Jesse L and Prince, Jacob S and Dowdle, Logan T and Nau, Matthias and Caron, Brad and Pestilli, Franco and Charest, Ian and others},
  journal={Nature neuroscience},
  volume={25},
  number={1},
  pages={116--126},
  year={2022},
  publisher={Nature Publishing Group US New York}
}

@article{kohler2016representation,
  title={Representation of maximally regular textures in human visual cortex},
  author={Kohler, Peter J and Clarke, Alasdair and Yakovleva, Alexandra and Liu, Yanxi and Norcia, Anthony M},
  journal={The Journal of Neuroscience},
  volume={36},
  number={3},
  pages={714--729},
  year={2016},
  publisher={Society for Neuroscience}
}

@article{charest2018glmdenoise,
  title={GLMdenoise improves multivariate pattern analysis of fMRI data},
  author={Charest, Ian and Kriegeskorte, Nikolaus and Kay, Kendrick N},
  journal={NeuroImage},
  volume={183},
  pages={606--616},
  year={2018},
  publisher={Elsevier}
}

@article{hubel1962receptive,
  title={Receptive fields, binocular interaction and functional architecture in the cat's visual cortex},
  author={Hubel, David H and Wiesel, Torsten N},
  journal={The Journal of Physiology},
  volume={160},
  number={1},
  pages={106--154},
  year={1962},
  publisher={Wiley Online Library},
  doi={10.1113/jphysiol.1962.sp006837},
  pmid={14449617}
}

@article{hubel1968receptive,
  title={Receptive fields and functional architecture of monkey striate cortex},
  author={Hubel, David H and Wiesel, Torsten N},
  journal={The Journal of physiology},
  volume={195},
  number={1},
  pages={215--243},
  year={1968},
  publisher={Wiley Online Library}
}

@article{stigliani2015temporal,
  title={Temporal processing capacity in high-level visual cortex is domain specific},
  author={Stigliani, Anthony and Weiner, Kevin S and Grill-Spector, Kalanit},
  journal={Journal of Neuroscience},
  volume={35},
  number={36},
  pages={12412--12424},
  year={2015},
  publisher={Society for Neuroscience}
}

@article{rokem2020fractional,
  title={Fractional ridge regression: a fast, interpretable reparameterization of ridge regression},
  author={Rokem, Ariel and Kay, Kendrick},
  journal={GigaScience},
  volume={9},
  number={12},
  pages={giaa133},
  year={2020},
  publisher={Oxford University Press}
}

@article{kay2013glmdenoise,
  title={GLMdenoise: a fast, automated technique for denoising task-based fMRI data},
  author={Kay, Kendrick N and Rokem, Ariel and Winawer, Jonathan and Dougherty, Robert F and Wandell, Brian A},
  journal={Frontiers in neuroscience},
  volume={7},
  pages={247},
  year={2013},
  publisher={Frontiers Media SA}
}

@article{luo2023brain,
  title={Brain diffusion for visual exploration: Cortical discovery using large scale generative models},
  author={Luo, Andrew and Henderson, Maggie and Wehbe, Leila and Tarr, Michael},
  journal={Advances in Neural Information Processing Systems},
  volume={36},
  pages={75740--75781},
  year={2023}
}

@article{prince2022improving,
  title={Improving the accuracy of single-trial fMRI response estimates using GLMsingle},
  author={Prince, Jacob S and Charest, Ian and Kurzawski, Jan W and Pyles, John A and Tarr, Michael J and Kay, Kendrick N},
  journal={Elife},
  volume={11},
  pages={e77599},
  year={2022},
  publisher={eLife Sciences Publications Limited}
}

@article{li2023blipdiff,
  title={Blip-diffusion: Pre-trained subject representation for controllable text-to-image generation and editing},
  author={Li, Dongxu and Li, Junnan and Hoi, Steven},
  journal={Advances in Neural Information Processing Systems},
  volume={36},
  pages={30146--30166},
  year={2023}
}

@InProceedings{li2023blip,
  title = 	 {{BLIP}-2: Bootstrapping Language-Image Pre-training with Frozen Image Encoders and Large Language Models},
  author =       {Li, Junnan and Li, Dongxu and Savarese, Silvio and Hoi, Steven},
  booktitle = 	 {Proceedings of the 40th International Conference on Machine Learning},
  pages = 	 {19730--19742},
  year = 	 {2023},
  editor = 	 {Krause, Andreas and Brunskill, Emma and Cho, Kyunghyun and Engelhardt, Barbara and Sabato, Sivan and Scarlett, Jonathan},
  volume = 	 {202},
  series = 	 {Proceedings of Machine Learning Research},
  month = 	 {23--29 Jul},
  publisher =    {PMLR},
  url = 	 {https://proceedings.mlr.press/v202/li23q.html}
}

@article{luo2023brainscuba,
  title={Brainscuba: Fine-grained natural language captions of visual cortex selectivity},
  author={Luo, Andrew F and Henderson, Margaret M and Tarr, Michael J and Wehbe, Leila},
  journal={arXiv preprint arXiv:2310.04420},
  year={2023}
}

@article{grill2004human,
  title={The human visual cortex},
  author={Grill-Spector, Kalanit and Malach, Rafael},
  journal={Annu. Rev. Neurosci.},
  volume={27},
  number={1},
  pages={649--677},
  year={2004},
  publisher={Annual Reviews}
}

@article{sergent1992functional,
  title={Functional neuroanatomy of face and object processing: a positron emission tomography study},
  author={Sergent, Justine and Ohta, Shinsuke and Macdonald, Brennan},
  journal={Brain},
  volume={115},
  number={1},
  pages={15--36},
  year={1992},
  publisher={Oxford University Press}
}

@article{goodfellow2020generative,
  title={Generative adversarial networks},
  author={Goodfellow, Ian and Pouget-Abadie, Jean and Mirza, Mehdi and Xu, Bing and Warde-Farley, David and Ozair, Sherjil and Courville, Aaron and Bengio, Yoshua},
  journal={Communications of the ACM},
  volume={63},
  number={11},
  pages={139--144},
  year={2020},
  publisher={ACM New York, NY, USA}
}

@article{ho2022cascaded,
  title={Cascaded Diffusion Models for High Fidelity Image Generation.},
  author={Ho, Jonathan and Saharia, Chitwan and Chan, William and Fleet, David J and Norouzi, Mohammad and Salimans, Tim},
  journal={J. Mach. Learn. Res.},
  volume={23},
  number={47},
  pages={1--33},
  year={2022}
}

@article{song2020score,
  title={Score-based generative modeling through stochastic differential equations},
  author={Song, Yang and Sohl-Dickstein, Jascha and Kingma, Diederik P and Kumar, Abhishek and Ermon, Stefano and Poole, Ben},
  journal={arXiv preprint arXiv:2011.13456},
  year={2020}
}

@article{jain2023selectivity,
  title={Selectivity for food in human ventral visual cortex},
  author={Jain, Nidhi and Wang, Aria and Henderson, Margaret M and Lin, Ruogu and Prince, Jacob S and Tarr, Michael J and Wehbe, Leila},
  journal={Communications Biology},
  volume={6},
  number={1},
  pages={175},
  year={2023},
  publisher={Nature Publishing Group UK London}
}

@article{gifford2025silico,
  title={In silico discovery of representational relationships across visual cortex},
  author={Gifford, Alessandro T and Jastrz{\k{e}}bowska, Maya A and Singer, Johannes JD and Cichy, Radoslaw M},
  journal={Nature Human Behaviour},
  volume={9},
  number={10},
  pages={2079--2098},
  year={2025},
  publisher={Nature Publishing Group UK London}
}

@article{cohen2000visual,
  title={The visual word form area: spatial and temporal characterization of an initial stage of reading in normal subjects and posterior split-brain patients},
  author={Cohen, Laurent and Dehaene, Stanislas and Naccache, Lionel and Leh{\'e}ricy, St{\'e}phane and Dehaene-Lambertz, Ghislaine and H{\'e}naff, Marie-Anne and Michel, Fran{\c{c}}ois},
  journal={Brain},
  volume={123},
  number={2},
  pages={291--307},
  year={2000},
  publisher={Oxford University Press}
}

@article{khosla2022highly,
  title={A highly selective response to food in human visual cortex revealed by hypothesis-free voxel decomposition},
  author={Khosla, Meenakshi and Murty, N Apurva Ratan and Kanwisher, Nancy},
  journal={Current Biology},
  volume={32},
  number={19},
  pages={4159--4171},
  year={2022},
  publisher={Elsevier}
}

@article{deangelis1995receptive,
  title={Receptive-field dynamics in the central visual pathways},
  author={DeAngelis, Gregory C and Ohzawa, Izumi and Freeman, Ralph D},
  journal={Trends in neurosciences},
  volume={18},
  number={10},
  pages={451--458},
  year={1995},
  publisher={Elsevier Current Trends}
}

@article{gu2022neurogen,
  title={NeuroGen: activation optimized image synthesis for discovery neuroscience},
  author={Gu, Zijin and Jamison, Keith Wakefield and Khosla, Meenakshi and Allen, Emily J and Wu, Yihan and St-Yves, Ghislain and Naselaris, Thomas and Kay, Kendrick and Sabuncu, Mert R and Kuceyeski, Amy},
  journal={NeuroImage},
  volume={247},
  pages={118812},
  year={2022},
  publisher={Elsevier}
}

@article{ratan2021computational,
  title={Computational models of category-selective brain regions enable high-throughput tests of selectivity},
  author={Ratan Murty, N Apurva and Bashivan, Pouya and Abate, Alex and DiCarlo, James J and Kanwisher, Nancy},
  journal={Nature communications},
  volume={12},
  number={1},
  pages={5540},
  year={2021},
  publisher={Nature Publishing Group UK London}
}

@article{bashivan2019neural,
  title={Neural population control via deep image synthesis},
  author={Bashivan, Pouya and Kar, Kohitij and DiCarlo, James J},
  journal={Science},
  volume={364},
  number={6439},
  pages={eaav9436},
  year={2019},
  publisher={American Association for the Advancement of Science}
}

@article{walker2019inception,
  title={Inception loops discover what excites neurons most using deep predictive models},
  author={Walker, Edgar Y and Sinz, Fabian H and Cobos, Erick and Muhammad, Taliah and Froudarakis, Emmanouil and Fahey, Paul G and Ecker, Alexander S and Reimer, Jacob and Pitkow, Xaq and Tolias, Andreas S},
  journal={Nature neuroscience},
  volume={22},
  number={12},
  pages={2060--2065},
  year={2019},
  publisher={Nature Publishing Group US New York}
}

@article{ponce2019evolving,
  title={Evolving images for visual neurons using a deep generative network reveals coding principles and neuronal preferences},
  author={Ponce, Carlos R and Xiao, Will and Schade, Peter F and Hartmann, Till S and Kreiman, Gabriel and Livingstone, Margaret S},
  journal={Cell},
  volume={177},
  number={4},
  pages={999--1009},
  year={2019},
  publisher={Elsevier}
}

@article{chang2019bold5000,
  title={BOLD5000, a public fMRI dataset while viewing 5000 visual images},
  author={Chang, Nadine and Pyles, John A and Marcus, Austin and Gupta, Abhinav and Tarr, Michael J and Aminoff, Elissa M},
  journal={Scientific data},
  volume={6},
  number={1},
  pages={49},
  year={2019},
  publisher={Nature Publishing Group UK London}
}

@inproceedings{iclrhmc,
 author = {Chen, Haomiao and Jamison, Keith and Sabuncu, Mert and Kuceyeski, Amy},
 booktitle = {International Conference on Learning Representations},
 editor = {C. Vondrick and B. Hariharan and C. Raffel and L. Pinto and D. Yang and A. Faust},
 pages = {142312--142338},
 title = {Beyond Grid-Locked Voxels: Neural Response Functions for Continuous Brain Encoding},
 url = {https://proceedings.iclr.cc/paper_files/paper/2026/file/e624a43624008a903f1e17c5cbf02ed8-Paper-Conference.pdf},
 volume = {2026},
 year = {2026}
}

@misc{cherti2023openclip,
  author = {Cherti, Mehdi and Beaumont, Romain and Wightman, Ross and Wortsman, Mitchell and Ilharco, Gabriel and Gordon, Cade and Schuhmann, Christoph and Schmidt, Ludwig and Jitsev, Jenia},
  title = {Reproducible Scaling Laws for Contrastive Language-Image Learning},
  howpublished = {In Proceedings of the IEEE/CVF Conference on Computer Vision and Pattern Recognition (CVPR), pp. 2818--2829},
  year = {2023},
  url = {https://doi.org/10.1109/CVPR52729.2023.00276}
}

@article{gu2023human,
  title={Human brain responses are modulated when exposed to optimized natural images or synthetically generated images},
  author={Gu, Zijin and Jamison, Keith and Sabuncu, Mert R and Kuceyeski, Amy},
  journal={Communications Biology},
  volume={6},
  number={1},
  pages={1076},
  year={2023},
  publisher={Nature Publishing Group UK, London}
}

@article {jang2025amygdala,
	author = {Jang, Grace and Kragel, Philip A.},
	title = {Understanding Human Amygdala Function with Artificial Neural Networks},
	volume = {45},
	number = {18},
	elocation-id = {e1436242025},
	year = {2025},
	doi = {10.1523/JNEUROSCI.1436-24.2025},
	publisher = {Society for Neuroscience},
	issn = {0270-6474},
	URL = {https://www.jneurosci.org/content/45/18/e1436242025},
	eprint = {https://www.jneurosci.org/content/45/18/e1436242025.full.pdf},
	journal = {Journal of Neuroscience}
}

@article{bar2006top,
  title={Top-down facilitation of visual recognition},
  author={Bar, Moshe and Kassam, Karim S and Ghuman, Avniel Singh and Boshyan, Jasmine and Schmid, Annette M and Dale, Anders M and H{\"a}m{\"a}l{\"a}inen, Matti S and Marinkovic, Ksenija and Schacter, Daniel L and Rosen, Bruce R and others},
  journal={Proceedings of the national academy of sciences},
  volume={103},
  number={2},
  pages={449--454},
  year={2006},
  publisher={National Academy of Sciences}
}

@article{fischl2002whole,
  title={Whole brain segmentation: automated labeling of neuroanatomical structures in the human brain},
  author={Fischl, Bruce and Salat, David H and Busa, Evelina and Albert, Marilyn and Dieterich, Megan and Haselgrove, Christian and Van Der Kouwe, Andre and Killiany, Ron and Kennedy, David and Klaveness, Shuna and others},
  journal={Neuron},
  volume={33},
  number={3},
  pages={341--355},
  year={2002},
  publisher={Elsevier}
}

@article{hoffmann2024anatomy,
  title={Anatomy-aware and acquisition-agnostic joint registration with SynthMorph},
  author={Hoffmann, Malte and Hoopes, Andrew and Greve, Douglas N and Fischl, Bruce and Dalca, Adrian V},
  journal={Imaging Neuroscience},
  volume={2},
  pages={imag--2},
  year={2024},
  publisher={MIT Press, 255 Main Street, 9th Floor, Cambridge, Massachusetts 02142, USA~…}
}

@article{glasser2016multi,
  title={A multi-modal parcellation of human cerebral cortex},
  author={Glasser, Matthew F and Coalson, Timothy S and Robinson, Emma C and Hacker, Carl D and Harwell, John and Yacoub, Essa and Ugurbil, Kamil and Andersson, Jesper and Beckmann, Christian F and Jenkinson, Mark and others},
  journal={Nature},
  volume={536},
  number={7615},
  pages={171--178},
  year={2016},
  publisher={Nature Publishing Group UK, London}
}

@article{fischl2012freesurfer,
  title={FreeSurfer},
  author={Fischl, Bruce},
  journal={Neuroimage},
  volume={62},
  number={2},
  pages={774--781},
  year={2012},
  publisher={Elsevier}
}

@article{benjamini1995controlling,
  title={Controlling the false discovery rate: a practical and powerful approach to multiple testing},
  author={Benjamini, Yoav and Hochberg, Yosef},
  journal={Journal of the Royal statistical society: series B (Methodological)},
  volume={57},
  number={1},
  pages={289--300},
  year={1995},
  publisher={Wiley Online Library}
}

@article{rose2024concentration,
  title={A concentration of visual cortex-like neurons in prefrontal cortex},
  author={Rose, Olivia and Ponce, Carlos R},
  journal={Nature Communications},
  volume={15},
  number={1},
  pages={7002},
  year={2024},
  publisher={Nature Publishing Group UK, London}
}

@article{wang2022tuning,
  title={Tuning landscapes of the ventral stream},
  author={Wang, Binxu and Ponce, Carlos R},
  journal={Cell reports},
  volume={41},
  number={6},
  year={2022},
  publisher={Elsevier}
}

@InProceedings{radford2021clip,
  title = 	 {Learning Transferable Visual Models From Natural Language Supervision},
  author =       {Radford, Alec and Kim, Jong Wook and Hallacy, Chris and Ramesh, Aditya and Goh, Gabriel and Agarwal, Sandhini and Sastry, Girish and Askell, Amanda and Mishkin, Pamela and Clark, Jack and Krueger, Gretchen and Sutskever, Ilya},
  booktitle = 	 {Proceedings of the 38th International Conference on Machine Learning},
  pages = 	 {8748--8763},
  year = 	 {2021},
  editor = 	 {Meila, Marina and Zhang, Tong},
  volume = 	 {139},
  series = 	 {Proceedings of Machine Learning Research},
  month = 	 {18--24 Jul},
  publisher =    {PMLR},
  url = 	 {https://proceedings.mlr.press/v139/radford21a.html},
}




\bigskip

\newpage

\begin{center}
    \Large\textbf{Supplementary Information}
\end{center}

\newcounter{supplementarynote}
\newcommand{\supplementarynote}[2]{%
    \refstepcounter{supplementarynote}%
    \section*{Supplementary Note \thesupplementarynote. #1}%
    \label{#2}%
}

\supplementarynote{Neural encoding model performance}{note:encoding-performance}

\begin{figure*}[!tp]
    \centering
    \includegraphics[width=\linewidth]{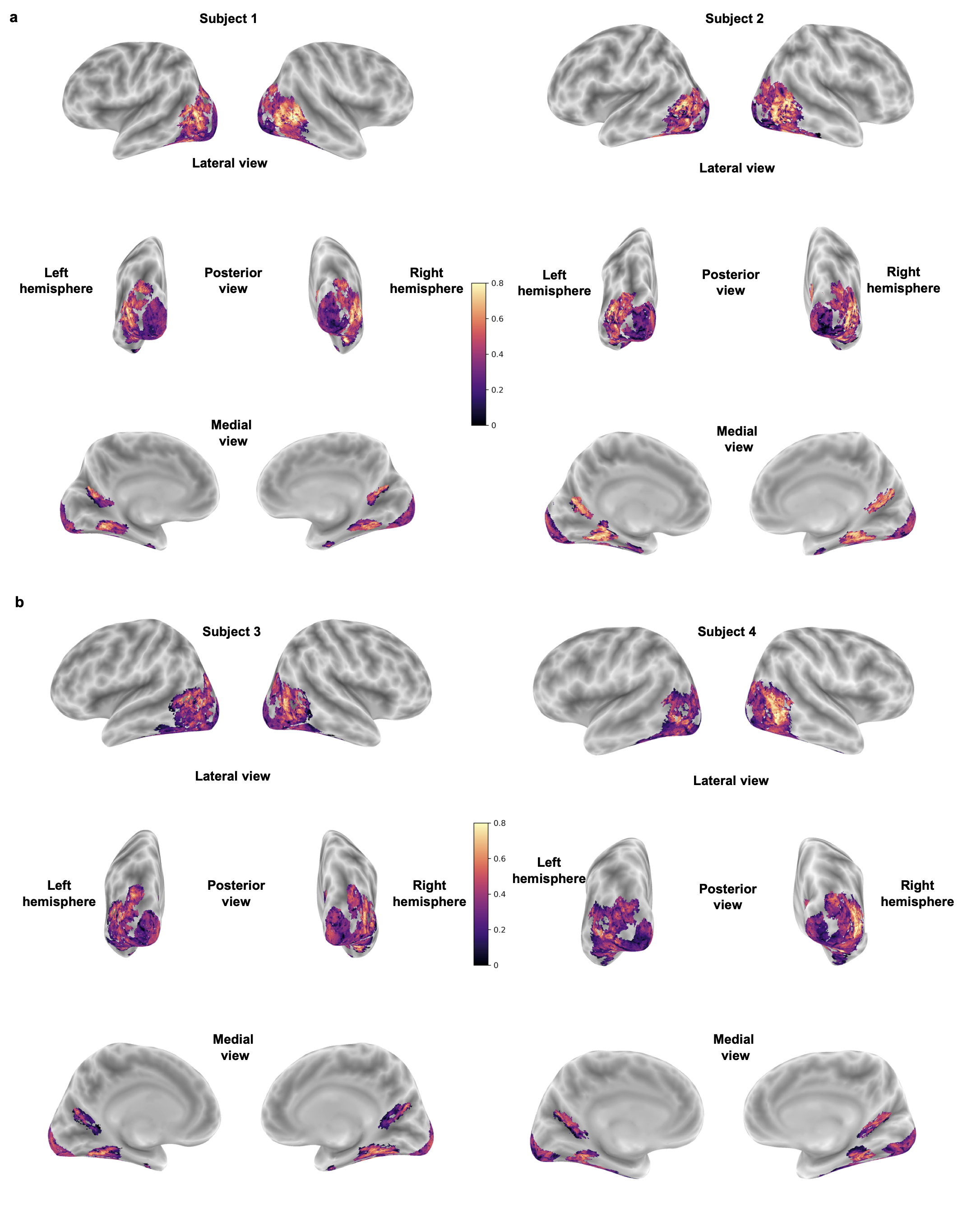}
    \caption{\textbf{BLIP-based optimization-encoder accuracy for NSD participants S1 - S4.}
    Maps show voxelwise Pearson correlations between predicted and measured responses
    across the shared NSD evaluation images. Only voxels included in the
    participant-specific visual-cortex model are displayed.}
    \label{fig:acc_1}
\end{figure*}

\begin{figure*}[!tp]
    \centering
    \includegraphics[width=\linewidth]{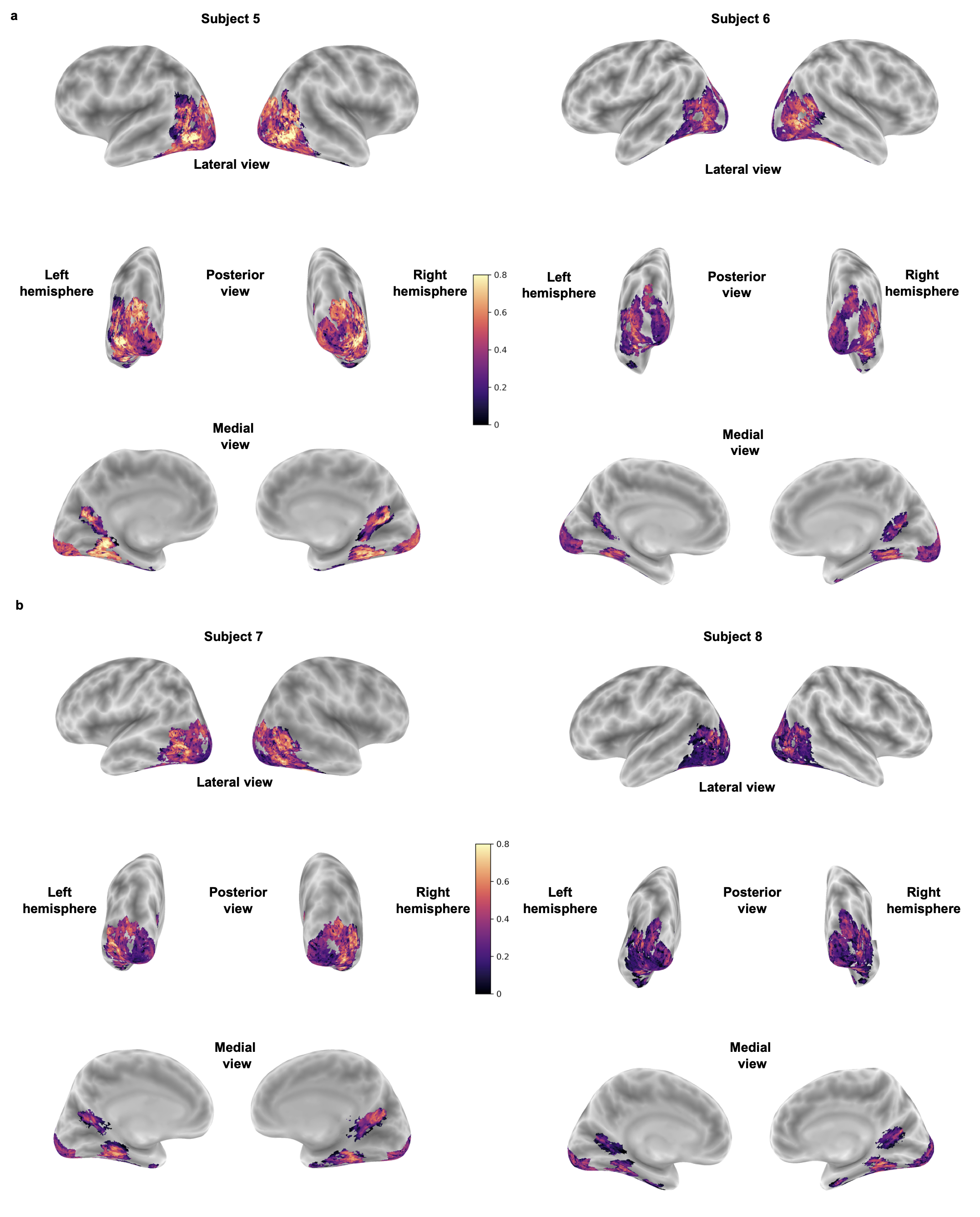}
    \caption{\textbf{BLIP-based optimization-encoder accuracy for NSD participants S5 - S8.}
    Maps show voxelwise Pearson correlations between predicted and measured responses
    across the shared NSD evaluation images. Only voxels included in the
    participant-specific visual-cortex model are displayed.}
    \label{fig:acc_2}
\end{figure*}

Encoding model accuracy was quantified as the Pearson correlation ($r$) between predicted and measured responses across the approximately
1,000 NSD images shared across participants and reserved for model evaluation (shared 1K).
For the visual-cortex models, accuracy was calculated for each voxel in the
participant-specific visual-cortex mask. Supplementary Figs.~\ref{fig:acc_1}--\ref{fig:acc_2}
show the accuracy of the BLIP-based optimization encoders for all eight NSD
participants. Voxels outside the modeled visual-cortex mask were not evaluated.

\newpage

We additionally evaluated the CLIP-based validation encoders for the visual
cortex and the BLIP- and CLIP-based encoders used to predict left and right
amygdala responses. The CLIP-based encoder was not used during Q-Former
embedding optimization. Performance for these models is shown in Supplementary Figs.~\ref{fig:regional_acc}.

\begin{figure*}[!tp]
    \centering
    \includegraphics[width=\linewidth]{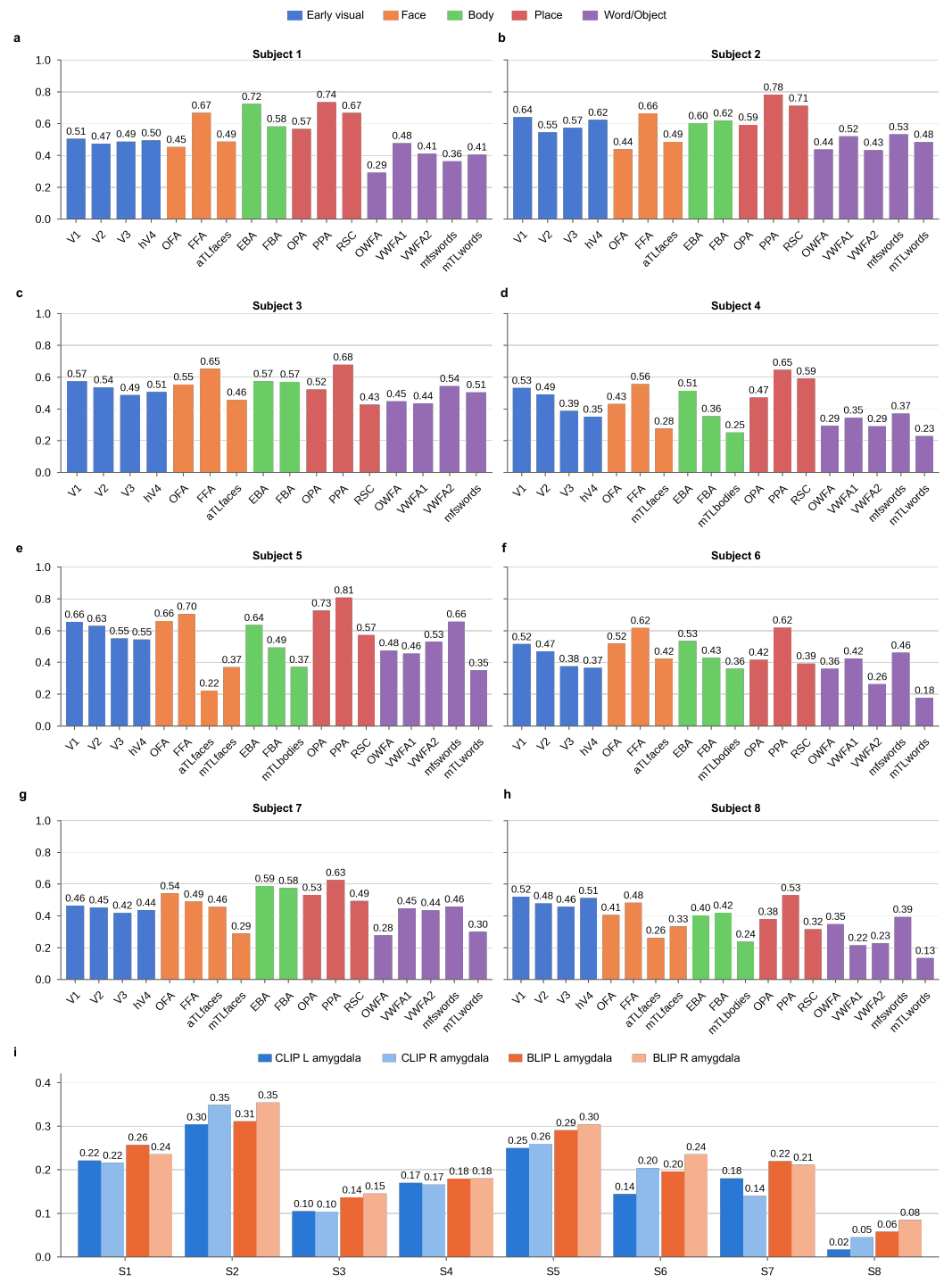}
    \caption{\textbf{Regional encoding-model accuracy in the visual cortex and amygdala.}Bars show region-level Pearson correlations between predicted and measured responses to the "shared 1K" evaluation images. CLIP-based validation-encoder accuracy is shown for each participant's visual regions and left and right amygdala. BLIP-based optimization-encoder accuracy is also shown for the left and right amygdala and mirrors the CLIP-based results.}
    \label{fig:regional_acc}
\end{figure*}

\FloatBarrier

\supplementarynote{Additional single-region optimization results}{note:single-region}

We repeated the single-region maximization analysis for all eight NSD
participants. For each participant, the same four initialization seed images were
used to optimize maximal activity responses in FFA, EBA, PPA, mfs-words, V1 and hV4. The figures show the
initialization images and images decoded from embeddings at 20\%, 50\%, 70\%
and 100\% of the total increase in the predicted target-region response. These
examples extend the participant S5 results presented in the main text, and provide even further support the idea of individually-specific optimal stimuli.

\begin{figure*}[!tp]
    \centering
    \includegraphics[width=\linewidth]{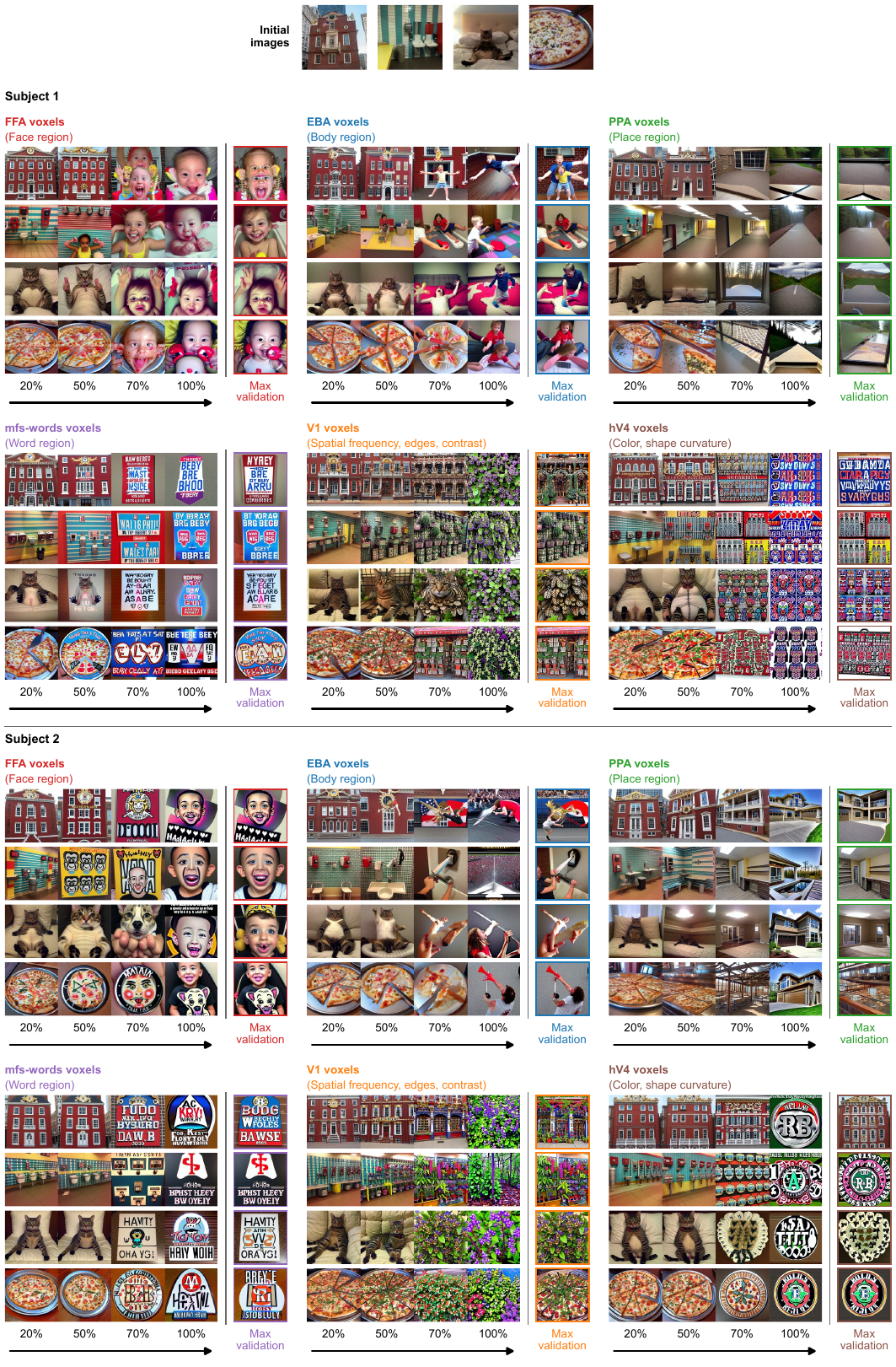}
    \caption{\textbf{NeuroVolve's single-region maximization results for NSD participants S1 and S2.} Images were generated from four common initialization images by maximizing the predicted response of FFA, EBA, PPA, mfs-words, V1 or hV4. Columns show images decoded at 20\%, 50\%, 70\% and 100\% of the total increase in the BLIP-based optimization objective, with the final column being the image that had the maximum value in the CLIP-based validation encoder.}
    \label{fig:supp_s1}
\end{figure*}

\begin{figure*}[!tp]
    \centering
    \includegraphics[width=\linewidth]{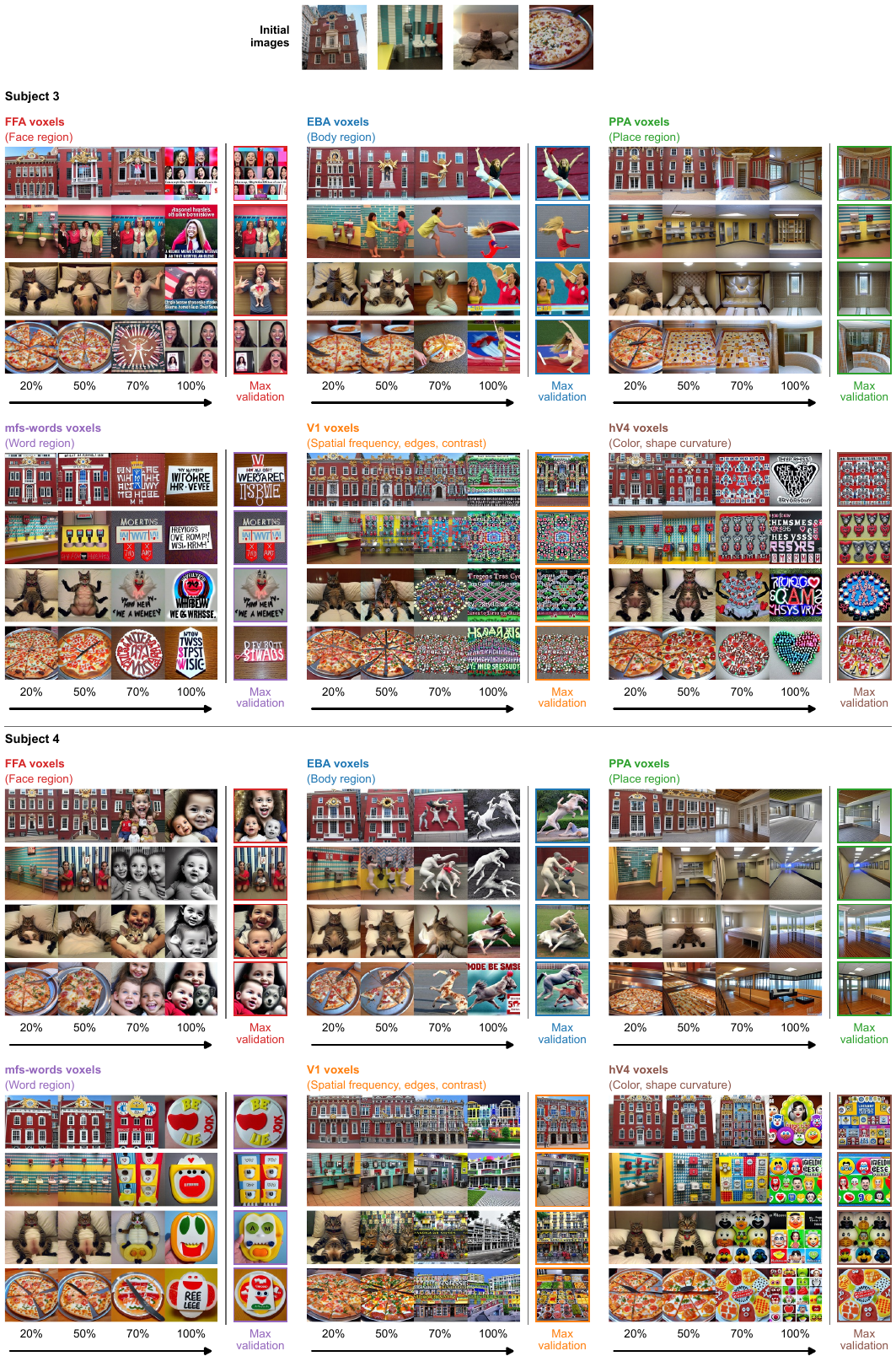}
    \caption{\textbf{Single-region image evolution for NSD participants S3 and S4.} Images were generated from four common initialization images by maximizing the predicted response of FFA, EBA, PPA, mfs-words, V1 or hV4. Columns show images decoded at 20\%, 50\%, 70\% and 100\% of the total increase in the BLIP-based optimization objective, with the final column being the image that had the maximum value in the CLIP-based validation encoder.}
    \label{fig:supp_s2}
\end{figure*}

\begin{figure*}[!tp]
    \centering
    \includegraphics[width=\linewidth]{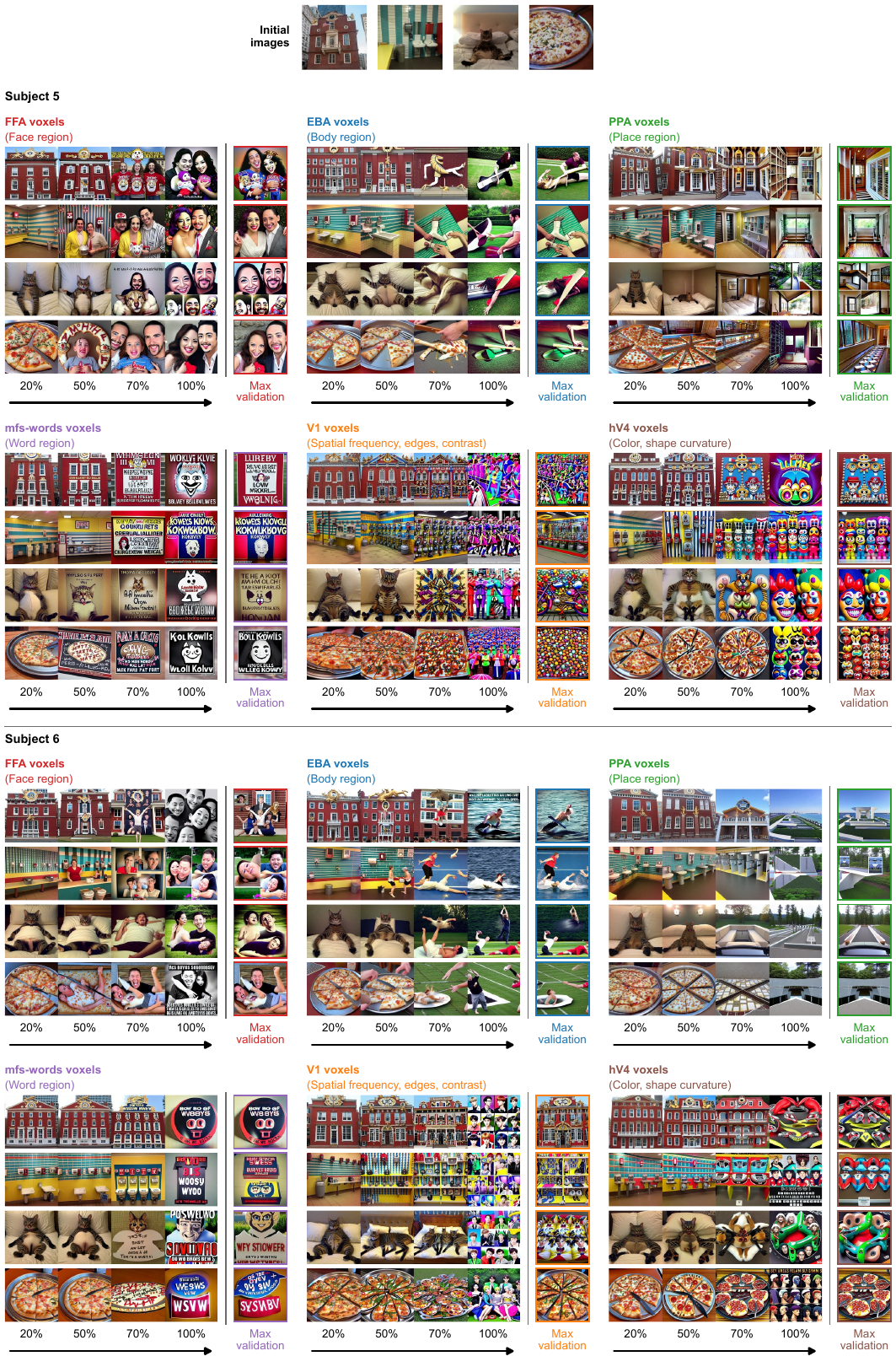}
    \caption{\textbf{Single-region image evolution for NSD participants S5 and S6.}Images were generated from four common initialization images by maximizing the predicted response of FFA, EBA, PPA, mfs-words, V1 or hV4. Columns show images decoded at 20\%, 50\%, 70\% and 100\% of the total increase in the BLIP-based optimization objective, with the final column being the image that had the maximum value in the CLIP-based validation encoder.}
    \label{fig:supp_s3}
\end{figure*}

\begin{figure*}[!tp]
    \centering
    \includegraphics[width=\linewidth]{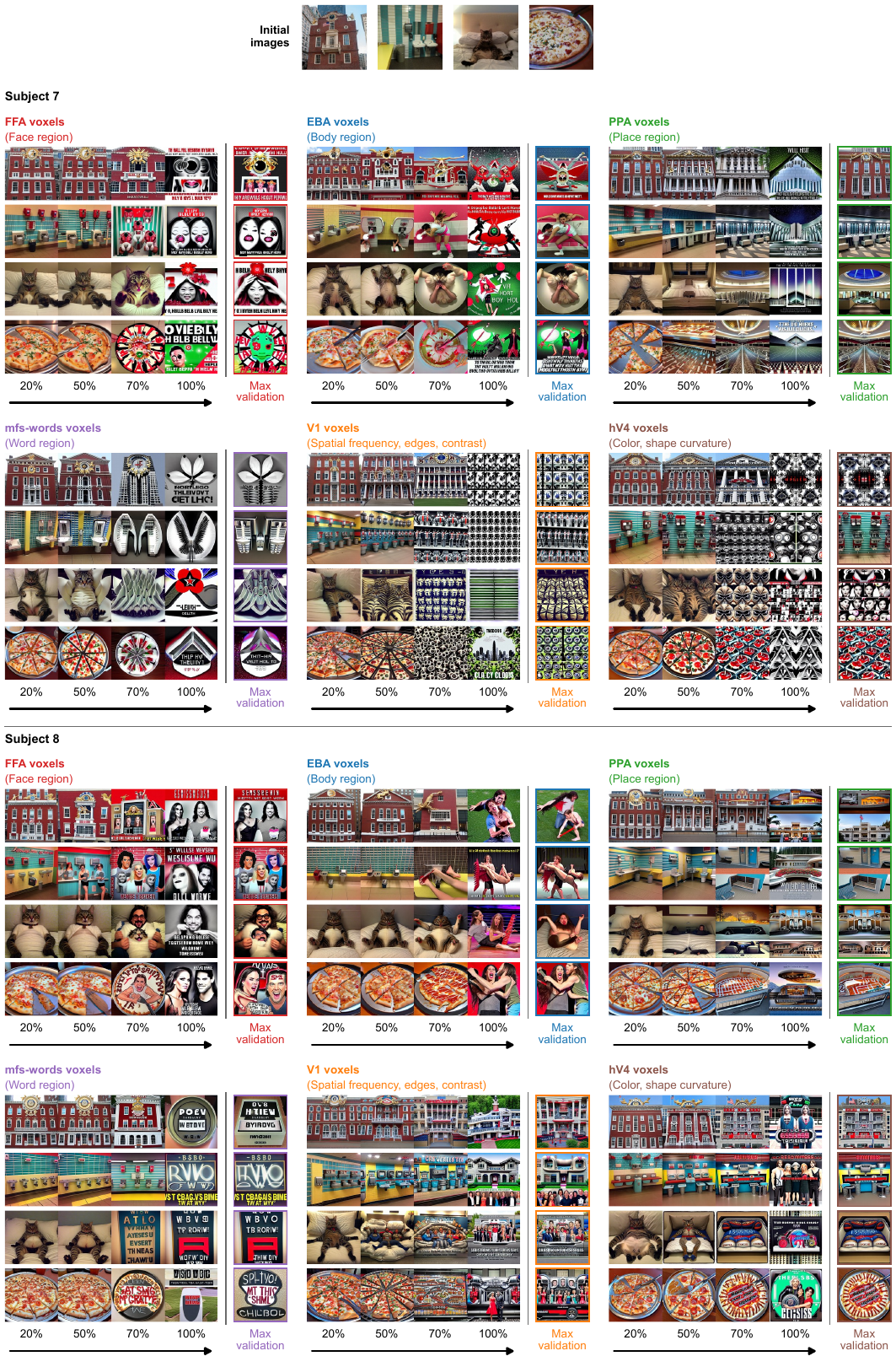}
    \caption{\textbf{Single-region image evolution for NSD participants S7 and S8.} Images were generated from four common initialization images by maximizing the predicted response of FFA, EBA, PPA, mfs-words, V1 or hV4. Columns show images decoded at 20\%, 50\%, 70\% and 100\% of the total increase in the BLIP-based optimization objective, with the final column being the image that had the maximum value in the CLIP-based validation encoder.}
   
    \label{fig:supp_s4}
\end{figure*}

\FloatBarrier

\supplementarynote{Semantic specificity of natural and NeuroVolve images}{note:semantic-specificity}

\paragraph{Image sets.}
We assessed whether NeuroVolve images contained visual content associated with
the targeted category-selective regions: FFA (faces), EBA (bodies), PPA (scenes)
and mfs-words (text). For each NSD participant and target ROI, we generated
1,000 images by maximizing the mean predicted response across voxels within
that ROI, starting from natural-image initializations.
The participant-specific CLIP-based validation encoder ranked the generated
images, and the 100 highest-ranked images were retained. Two natural-image
reference sets were constructed for each participant and target. For the
measured-response reference, we selected the 100 images with the highest measured
mean response from the NSD images actually viewed by that participant. For the
predicted-response reference, we scored the full pool of 10,000 available natural
images using that participant's validation encoder and selected the 100 images
with the highest predicted mean response.

\paragraph{Zero-shot semantic classification.}
We classified each image into one of four categories---face, body, scene or
text---using the frozen CoCa ViT-L/14 model. This model was distinct from the
BLIP-based optimization encoder and the CLIP ViT-H/14 validation encoder and
was not used for neural encoding, embedding optimization or image selection.

All category-specific prompts (Table~\ref{table:semantic_prompts}) were encoded
once using the CoCa text encoder. Images were processed using the model's
standard preprocessing transform and encoded using its image encoder. Image
and text embeddings were normalized to unit Euclidean norm, and image--prompt
similarities were calculated as their cosine similarities multiplied by 100.
Each category's score was the highest similarity across its prompts
(max-pooling), allowing an image to match any one description of that category.
The four category scores were converted to probabilities using a softmax, and
the image was assigned to the category with the highest probability. This
assignment is equivalent to selecting the category with the highest max-pooled
similarity. No confidence threshold or rejection criterion was applied, so
every image received exactly one label.

Semantic specificity was the percentage of images assigned to the expected
target category: face for FFA, body for EBA, scene for PPA and text for
mfs-words. Images assigned to a category other than the target category counted as
non-specific. The same procedure was applied to all three top-100 image sets
for every participant and target region.

\paragraph{Semantic specificity of image sets across participants.}
Table~\ref{table:categ-clip} summarizes semantic specificity for the three
image sets as mean $\pm$ s.d. across the eight NSD participants.

\begin{table}[!htbp]
\centering
\resizebox{\linewidth}{!}{%
\begin{tabular}{lcccc}
\toprule
Image set
& FFA (face)
& EBA (body)
& PPA (scene)
& mfs-words (text) \\
\midrule
NeuroVolve top 100 (validation prediction)
& $\mathbf{84.8\pm14.2}$
& $\mathbf{70.9\pm25.5}$
& $\mathbf{97.9\pm6.0}$
& $\mathbf{76.8\pm27.9}$ \\
NSD top 100 (validation prediction)
& $60.0\pm8.9$
& $69.9\pm9.6$
& $90.9\pm10.1$
& $50.0\pm12.1$ \\
NSD top 100 (measured response)
& $45.5\pm7.2$
& $68.3\pm6.4$
& $73.0\pm13.4$
& $34.6\pm11.2$ \\
\bottomrule
\end{tabular}%
}
\caption{\textbf{Semantic specificity across NSD participants.}
Values are the percentages of images assigned to the category associated with
the target ROI by zero-shot CoCa ViT-L/14 classification, reported as
mean $\pm$ s.d. across eight participants. The measured-response natural reference
contains the 100 highest-response images among those viewed by each participant.
The predicted-response natural reference contains the 100 highest-ranked images
from the full pool of 10,000 available natural images, scored by the
participant-specific CLIP-based validation encoder. NeuroVolve sets contain
the 100 generated images ranked highest by the same validation encoder.}
\label{table:categ-clip}
\end{table}

\paragraph{Classification prompts.}
The prompt wording is reproduced below without modification. The category
labels ``scene'' and ``text'' correspond to the place- and word-selective
targets, respectively.

\begin{table}[!htbp]
\centering
\begin{tabular}{@{}lp{0.82\linewidth}@{}}
\toprule
Category & Prompts \\
\midrule
Face &
A face facing the camera; A photo of a face; A photo of a human face;
A photo of faces; A photo of a person's face; A person looking at the camera;
People looking at the camera; A portrait of a person; A portrait photo \\
\addlinespace
Body &
A photo of a torso; A photo of torsos; A photo of limbs; A photo of bodies;
A photo of a person; A photo of people \\
\addlinespace
Scene &
A photo of a bedroom; A photo of an office; A photo of a hallway;
A photo of a doorway; A photo of interior design; A photo of a building;
A photo of a house; A photo of nature; A photo of landscape; A landscape photo;
A photo of trees; A photo of grass \\
\addlinespace
Text &
A photo of words; A photo of glyphs; A photo of a glyph; A photo of text;
A photo of numbers; A photo of a letter; A photo of letters; A photo of writing;
A photo of text on an object \\
\bottomrule
\end{tabular}
\caption{\textbf{Text prompts for zero-shot semantic classification.}
The same category-specific prompt sets were used to classify natural and
NeuroVolve images with CoCa ViT-L/14.}
\label{table:semantic_prompts}
\end{table}

\FloatBarrier

\supplementarynote{Encoder-predicted responses to natural and NeuroVolve images}{note:predicted-response}

We used the participant-specific CLIP-based validation encoders to compare
predicted brain responses to top 100 natural and NeuroVolve images chosen and created to maximize a single region's response. For each participant and
ROI, the same encoder was used to rank both image sources. We compared the 100
highest-ranked natural NSD images with the 100 highest-ranked NeuroVolve images.
Because the same validation encoder was used for ranking and evaluation, this
analysis provides an \emph{in silico} characterization of predicted response magnitude
rather than an independent statistical validation.

\newpage

\supplementarynote{Shared seed images across subjects' top-50 NeuroVolve image sets}{note:overlap}

All participants' NeuroVolve images for a region were optimized from the same 1,000 seed images in the shared 1K (Fig.~\ref{fig:subject_quant}). For each region and participant pair ($n=28$), we counted the initial seeds shared by the two participants' top 50 NeuroVolve image sets. If the top 50 were a random sample of the seeds, a pair would share 2.5 seeds on average (corresponding to Dice of 0.05). We compared the mean Dice coefficient across subject pairs with a null distribution from 100,000 simulations in which each participant's top 50 were drawn at random from the 1,000 seeds; one-tailed $p$ values were the proportion of simulations with a mean at least as large as observed. None of the region's overlap exceed chance (all uncorrected $p > 0.05$), indicating that number of shared seeds likely had minimal effect on the within- and across-subject similarity of NeuroVolve images (Fig.~\ref{fig:subject_quant}).

\newpage

\supplementarynote{Subject specificity relative to measured fMRI responses}{note:subject-specificity}

In the main analysis (Fig.~\ref{fig:subject_quant}), natural images were ranked using predicted responses to images from the validation encoder. We repeated the analysis in Fig.~\ref{fig:subject_quant} using natural images ranked by measured fMRI responses. However, because of the measurement noise, they provide a noisier estimate of image preference than encoding-model predictions.

For each image, voxelwise responses were first averaged across repeated presentations. For each subject and region, these repetition-averaged responses were then averaged across all voxels in the region. We used the resulting regional responses to select the 50 highest-responding images from the approximately 9,000 subject-unique natural images presented to that subject. We compared these image sets with the same top 50 NeuroVolve images used in the main analysis in Fig.~\ref{fig:subject_quant}. 

The results were consistent with the main findings.  For natural images selected using measured responses, within-subject similarity was numerically higher than across-subject similarity in the regions, but the difference was not statistically significant for all regions, likely due to the noise of the measured fMRI responses. In contrast, the top 50 NeuroVolve images showed significantly higher within-subject than across-subject similarity in all six regions (all Benjamini-Hochberg corrected p < 0.05). 

\begin{figure*}[!htb]
    \centering
    \begin{subfigure}[t]{\linewidth}
        \includegraphics[width=\linewidth]{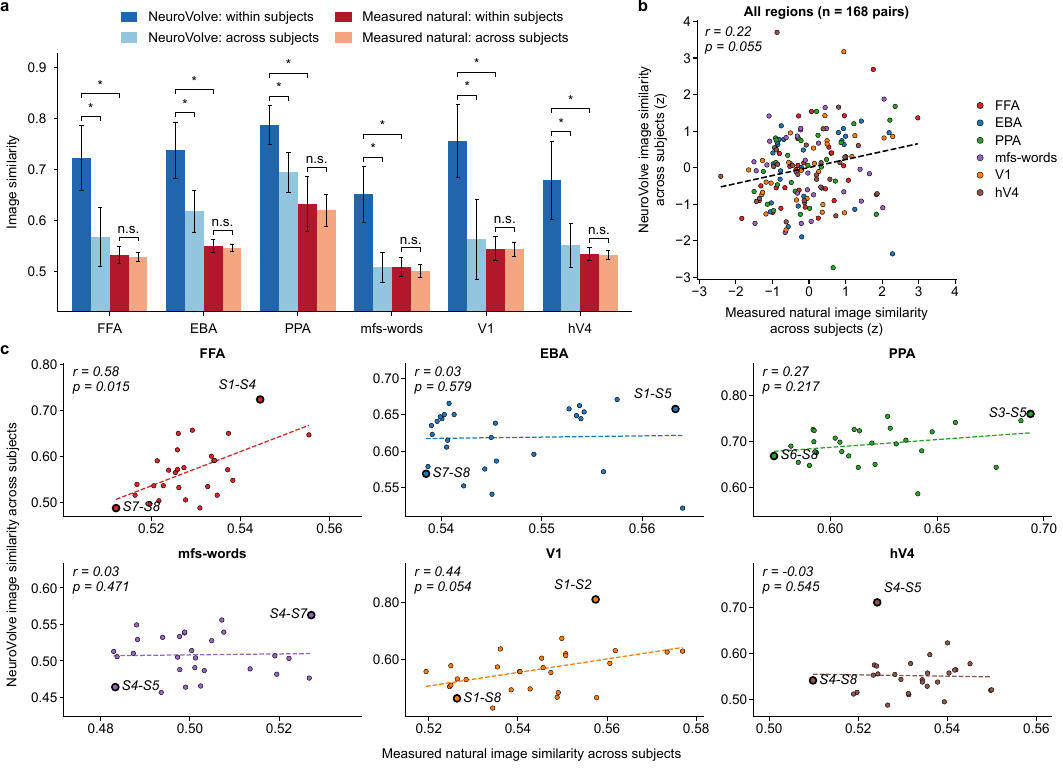}
    \end{subfigure} \caption{\textbf{NeuroVolve preserves within- and across-individual measured differences in optimal images.} \textbf{a}, Within- and across-subject similarity of the top 50 NeuroVolve images (blue) and the top 50 measured natural NSD images (red) for each target region, ranked by the region's measured activity in the NSD dataset. Similarity is the mean cosine similarity between image embeddings over all image pairs within or across subjects. Bars show the mean and error bars the s.d. across participants ($n=8$) or participant pairs ($n=28$). Brackets show two-tailed exact Wilcoxon signed-rank tests ($n=8$), with  values corrected across the 18 tests in \textbf{a} by the Benjamini--Hochberg false discovery rate procedure; *$q<0.05$; n.s., not significant. \textbf{b}, Across-subject similarity of NeuroVolve versus measured natural images for all subject pairs and regions ($n=168$; 28 pairs $\times$ 6 regions), $z$-scored within each region. The dashed line shows the least-squares fit, and $r$ is the pooled Pearson correlation. \textbf{c}, The same data as in \textbf{b}, shown for each region on the original similarity scale ($n=28$ pairs per panel).}
        \label{fig:meas_subject}
\end{figure*}

Then we tested if two individuals had more similar top 50 measured natural images they also had more similar top 50 NeuroVolve images. Across regions, the across-subject similarity of the NeuroVolve and natural-image sets was positively associated, although the relationship did not reach significance ($r = 0.22, p = 0.055$, after z-scoring within each region; Supplementary Fig.~\ref{fig:meas_subject}b). Region-specific analyses revealed a significant association in FFA ($r = 0.58, p = 0.015$) and a similar but nonsignificant trend in V1 ($r = 0.44, p = 0.054$). Associations were weaker in EBA, PPA, mfs-words, and hV4 (Supplementary Fig.~\ref{fig:meas_subject}c).

These results indicate that the subject-pair structure captured by NeuroVolve in FFA was also present when natural images were selected directly using measured fMRI responses. The associations in V1 and across regions were in the same direction but did not reach statistical significance. Because measurement noise reduces the reliability of rankings based on observed fMRI responses, we consider the analysis based on validation-encoder predictions to provide a more reliable estimate of subject-specific image preferences.

\clearpage

\newpage
\supplementarynote{Individual-level amygdala activation analysis}{note:pooled_amygdala}

The main amygdala analyses (Fig.~\ref{fig:amygdala_cortical}) averaged image responses across participants to increase signal-to-noise ratios before testing for group differences. To account for possible subject variability (but with the disadvantage of decreased SNR in the individual measures), we also replicated the analyses on every participant's response to every image (6 participants $\times$ 325 images, with individual-level responses averaged over the two repeated image viewings) using a linear mixed-effects model:
\begin{equation}
  \text{brain response} \sim \text{condition} + (1 \mid \text{participant}) + (1 \mid \text{image})
        + (1 \mid \text{participant}:\text{condition}).
\end{equation}
The random effects give each participant their own baseline and condition effect, and each image its own effect, so a condition effect is significant only if it holds across participants and images. Models were fit by maximum likelihood with statsmodels MixedLM and contrasts tested with Wald $z$ tests, (1) for the amygdala ROI in Fig.~\ref{fig:amygdala_cortical}a (Holm-corrected across the 10 condition pairs) and (2) separately for each of the 378 regions (360 HCP--MMP1 cortical parcels and 18 FreeSurfer subcortical structures) with FDR-correction across all regions). 

The results largely matched the main analysis (Supplementary Fig.~\ref{fig:amygdala_cortical}). NeuroVolve High and Low images separated amygdala responses as natural images did, with NeuroVolve High exceeding Natural Zero ($p = 0.034$) and NeuroVolve Low ($p = 3.3 \times 10^{-4}$). Across regions, NeuroVolve and natural images again produced similar patterns of contrast to the Natural Zero images (High $r = 0.84$, $p = 8.0 \times 10^{-100}$; Low $r = 0.68$, $p = 2.6 \times 10^{-52}$). In the right amygdala, both Natural High and NeuroVolve High responses exceeded Natural Zero (FDR $q = 0.001$ and $q = 0.016$). The left amygdala showed larger effects in the same direction that did not survive correction ($q = 0.17$ and $q = 0.25$), reflecting the greater variability across individual participants. Low images reduced amygdala responses relative to Natural Zero in both hemispheres, but none of these differences survived correction ($q \geq 0.15$). Differences between NeuroVolve and natural images had the same pattern of differences as in the main analysis, though fewer survived correction.

\FloatBarrier

\end{document}